\documentclass[11pt]{article}

\usepackage{longtable}
\usepackage{array}
\usepackage{booktabs}
\usepackage[utf8]{inputenc}
\usepackage[T1]{fontenc}
\usepackage{tipa}
\usepackage{booktabs}
\usepackage{amsmath}
\usepackage{pdflscape}
\newcommand{\ipa}[1]{\textipa{#1}}
\newcommand{\lang}[1]{\texttt{#1}}
\usepackage{subcaption}
\usepackage{graphicx}
\usepackage[T3,T1]{fontenc} 
\usepackage{tipa} 
\usepackage{longtable}

\usepackage[preprint]{acl}

\usepackage{times}
\usepackage{latexsym}

\usepackage[T1]{fontenc}

\usepackage[utf8]{inputenc}

\usepackage{microtype}

\usepackage{inconsolata}

\usepackage{graphicx}

\title{Evaluating Bias in Phoneme-Based Automatic Speech Recognition Systems: An Analysis of IPA Transcription Models}

\author{
  Maneesha Rani Saha \thanks{These authors contributed equally.} \\
  University of Utah\\
  \texttt{maneesha.saha@utah.edu} 
    \\\And
  Catherine Bao \footnotemark[1] \\
  University of Utah \\
  \texttt{u1459030@utah.edu} 
  \\\And
  Neal Patwari \\
  University of Utah \\
  \texttt{neal.patwari@utah.edu} \\
  }

\usepackage{tipa} 
\usepackage{textcomp} 

\begin{document}
\maketitle
\begin{abstract}
As automatic speech recognition (ASR) systems shift toward multilingual support and low-resource language modeling, phoneme-based layers serve as a critical language-agnostic foundation. 
However, most evaluations of ASR's demographic biases related to race, age, gender, and accent focus on standard grapheme-based ASR systems with comparatively little emphasis on phoneme-based systems.
In this study, we evaluate the performance of  WhisperIPA and ZIPA, two state-of-the-art open-source systems that generate International Phonetic Alphabet (IPA) transcriptions. Our evaluation includes existing multilingual speech corpora and demographically annotated English-language corpora,
comparing model-generated IPA transcriptions against grapheme-to-phoneme (G2P) systems using both standard phoneme error rate (PER) and a proposed Soft PER metric that tolerates linguistically similar phoneme substitutions. Our analysis examines how performance varies across language, gender, accent, ethnicity, and age, revealing persistent disparities even after accounting for acceptable phonemic variation. 
These findings, while limited, provide insight into potential sources of bias and inform the development of more inclusive and linguistically robust phoneme-based ASR systems. Our code and data are publicly available.\footnote{https://github.com/Maneesha28/ipa-asr-transcription-bias-evaluation}

\end{abstract}

\section{Introduction}

Automatic Speech Recognition (ASR) systems are a foundational component of modern agent-based technology. Current ASR systems typically rely on grapheme representations: graphemes are written units of language (e.g., letters or characters), while phonemes are units of sound that represent how words are pronounced. Existing research shows that grapheme-based ASR systems exhibit systematic disparities across demographic groups, including race, gender, age, accent, and language background \cite{tatman2017effects, tatman2017gender, dichristofano2023performance, jahan2025unveiling, elghazaly2025exploring, scott2025sociophonetic, serditova2025automatic, cunningham2025toward}. ASR evaluation currently mostly focuses on word error rates (WER) \cite{feng2021quantifying}.

While a majority of industry models remain grapheme-based, there has been a growing trend toward integrating IPA layers for multilingual capability \cite{li2020universal, xu2021simple}. The International Phonetic Alphabet (IPA) captures fine-grained phonetic details and cross-linguistic sound variations. Multilingual training on IPA tokens reduces phonetic token error rate by 60–70\% in high-resource languages and 14–42\% in low-resource languages \cite{zelasko2022discoveringphoneticinventoriescrosslingual}, and including IPA layers enables more effective generalization by providing a more direct mapping between the acoustic signal and its phonological categories. This makes recent work in phonetic \cite{bharadwaj2026prismbenchmarkingphonerealization, goriely2025ipa} and IPA models \cite{fang2020using, taguchi2023universalautomaticphonetictranscription, zhu-etal-2024-taste, lee2025leveraging, zhu2025zipa}, integrated as standalone or intermediate layers, valuable for multilingual and low-resource ASR systems.

Despite this growing importance, bias research has focused almost exclusively on grapheme-based, WER-measured systems; comparatively little has been done on phoneme representations. This gap matters in both directions: if disparities are already present at the phoneme level, they may propagate through downstream components of integrated ASR systems; conversely, if phonetic representations prove more robust, they may lead to more equitable speech recognition. Phonemic-level demographic bias thus remains largely underexplored, motivating the present study.

Since the same phoneme can be realized differently across accents, dialects, and languages, strict Phonetic Error Rate (PER) may penalize valid pronunciation variation, introducing noise when evaluating demographic disparities. To reduce this effect,  we introduce a soft variant of PER that tolerates minor phonemic variation. 
We evaluate how this PER variant changes phoneme ASR bias analysis. Our contributions include:
\begin{itemize}
    \item[1.] Introducing and evaluating a Soft PER that tolerates minor phonemic variation rather than strictly penalizing all substitutions. We compare Soft PER with standard PER to determine whether it yields a more robust metric for evaluating demographic bias.
    \item[2.] Evaluating the biases of IPA-based phoneme ASR models across accent, gender, age, ethnicity and language. Our evaluation spans 11 languages (English, Hindi, Bangla, Panjabi, Tamil, Telugu, Arabic, French, Spanish, Turkish, and Shona). We also analyze different age groups, gender, four ethnicity groups, and English accents across eight accent/dialect groups.
\end{itemize}
\section{Related Work}
End-to-end ASR systems have increased accuracy by scaling data and model size, especially with transformer‑family architectures \cite{radford2022whisper}. Large models such as wav2vec 2.0 \cite{baevski2020wav2vec}, Whisper \cite{radford2022whisper}, and XLS-r \cite{babu22_interspeech} demonstrate that grapheme-based models can generalize to dozens of languages. Recent work focuses on even larger multilingual models that replace language‑specific systems and target lower-resource languages \cite{meng2025dolphin, omnilingual2025omnilingual, ramirez2024anatomyindustrialscalemultilingual}.

Despite these advances, ASR is concentrated in high-resource languages with standardized orthographic transcriptions \cite{gale-etal-2023-mixed-BORT, gorisch-schmidt-2024-evaluating-ortho}. Grapheme-based models rely on widely adopted writing systems in paired speech-text data. For languages and conversation without a formal writing system, non-standardized orthography, limited textual resources, or code-switching, grapheme models are less reliable or unavailable. Most grapheme mappings do not consider phonetic variation \cite{fang2020using, omnilingual2025omnilingual, ramirez2024anatomyindustrialscalemultilingual}.

By mapping acoustically similar sounds to shared phonetic symbols, these systems can transfer knowledge from high-resource to truly under-resourced languages \cite{daul-etal-2026-linguistically}. Currently, architectures introduce an explicit phonetic layer that improves robustness and data efficiency in low-resource settings \cite{lee2025leveraging}. Incorporating phonemic supervision can reduce the ultimate WER \cite{fu2025pac, scott2025sociophonetic}. 

Phonemic models are particularly advantageous for orthographic limitations or linguistic variability. Phonemic/phonetic token models outperform orthographic baselines in both data-constrained and cross-lingual domains \cite{daul-etal-2026-linguistically}. Shared phonetics allow easier cross-lingual transfer and generalize underrepresented languages \cite{goriely2025ipa, zhu2025zipa, devi2026ipaforassamesespeech}. Phonemic modeling directly captures acoustic distinctions, thus improving robustness to pronunciation variability \cite{lyu2008language, solorio2008learning, onda2026advancedmodelinginterlanguagespeech}.

Rather than replacing grapheme-based systems, phonetic layers are often integrated within them. Multi-task learning frameworks jointly predict phonemes and graphemes. Hybrid approaches have demonstrated improved robustness in noisy, low-resource, and multilingual settings, suggesting that phonetic supervision can complement larger ASR architectures \cite{fang2020using}. 

Bias in ASR systems has been extensively studied in grapheme-based systems, with prior research showing significant disparities in WER across demographic groups --- such as race, gender, dialect, and accent \cite{tatman2017effects, dichristofano2023performance, jahan2025unveiling}. An ASR system exhibits bias when its performance differs by user demographics \cite{zhang2023exploring}. Within grapheme-based models, commercial systems consistently have higher error rates for women compared to men \cite{ngueajio2022hey, aryal2023hey}. Models often have higher error rates for children \cite{bhardwaj2022automatic}. Non-native and non-urban accents have been associated with higher error rates \cite{serditova2025automatic, torgbi2025adapting, wassink2022uneven}, with second-language speech accents showing higher error rates than speakers of non-standard or regional accents \cite{feng2021quantifying, zhang2022mitigating}. For example, commercial models produce  twice as many errors for African American speakers as for white speakers of English \cite{koenecke2020racial, martin2020understanding}. 

Standardized evaluation benchmarks and bias audits have been introduced to quantify these performance gaps and assess fairness in widely used ASR models. In contrast, investigations into phonetic or IPA-based ASR models remain limited. As phoneme-level systems become more common, either as full speech transcription models or as intermediate components within larger architectures, it is increasingly important to understand how they may contribute to or mitigate existing biases. 
\section{Methods}
\subsection{Datasets}
\label{dataset}
We evaluate phoneme recognition models across two complementary dimensions: cross-lingual robustness and demographic variation. 

\paragraph{Cross-lingual Datasets.} To assess multilingual phoneme recognition performance, we use IPAPACK (FLEURS-IPA) \cite{zhu-etal-2024-taste}, MediaSpeech \cite{kolobov2021mediaspeechmultilanguageasrbenchmark} and WAXAL \cite{diack2026waxallargescalemultilingualafrican}. From these multilingual corpora, we select 11 languages for evaluation, including high-resource languages (English, Spanish, and French) and lower-resource languages (Hindi, Bangla, Panjabi, Tamil, Telugu, Arabic, Turkish, and Shona). Languages are selected based on support from both the G2P pipeline and the evaluated IPA models. Table \ref{appendix:dataset_overview} in Appendix \ref{appendix:dataset} provides more details about the language distributions of these datasets.

\paragraph{Demographic Datasets.} To analyze demographic bias and variation in phoneme recognition, we use English speech datasets with speaker metadata: the Corpus of Regional African American Language (CORAAL) \cite{kendall2018corpus}, the Edinburgh International Accents of English Corpus (EdAcc) \cite{sanabria2023edinburghinternationalaccentsenglish}, the Sonos Voice Control Bias Assessment Dataset (SVC) \cite{sekkat2024sonosvoicecontrolbias}, and WAXAL \cite{diack2026waxallargescalemultilingualafrican}. These datasets contain annotations for demographic attributes, including age, gender, ethnicity, accent, geographic region, and native-speaker status, with the demography presented in Table \ref{appendix:dataset_demographics} in Appendix \ref{appendix:dataset}.

\subsection{Models}

In this work, we evaluate two IPA-based ASR models which allows us to maximize coverage in cross-lingual evaluation, since they have the largest number of overlapping supported languages.

\paragraph{WhisperIPA.} We evaluate WhisperIPA\footnote{https://huggingface.co/neurlang/ipa-whisper-base}, a transformer-based encoder--decoder model built on the Whisper architecture \cite{radford2022whisper}. Whisper is originally pre-trained on 680,000 hours of weakly supervised multilingual speech data. The base WhisperIPA variant, with approximately 74M parameters, is further fine-tuned for direct speech-to-IPA transcription on 15,000 labeled synthetic IPA audio samples. The fine-tuning data is derived from Common Voice 21, which covers $>$70 languages with corresponding phonetic annotations.

\paragraph{ZIPA.} We evaluate ZIPA \cite{zhu2025zipa}, a family of efficient multilingual phone recognition models built on the Zipformer architecture \cite{yao2024zipformer}, that includes both transducer-based and CTC-based variants. In this work, we use the large consistency-regularized CTC model with noisy-student training. 
The ZIPA-CR-NS\footnote{https://huggingface.co/anyspeech/zipa-large-crctc-ns-800k} large variant has approximately 300M parameters and is trained on IPAPack++, a large-scale multilingual speech corpus with 17,132 hours of normalized phone transcriptions across 88 languages. The noisy-student training stage further incorporates approximately 11,000 hours of pseudo-labeled multilingual speech from $>$4,000 languages, improving cross-linguistic phone recognition performance.

\subsection{Experimental Setup}\label{sec:exp-setup}

All models are evaluated in a zero-shot setting without additional fine-tuning on the evaluation datasets in order to measure out-of-the-box phoneme recognition performance. After both models generate IPA transcriptions for every audio sample, outputs are post-processed to remove special tokens, normalize whitespace, and ensure consistency with the reference phoneme format. Output post-processing details are described in Appendix \ref{appendix:ipa-normalizer}.

Reference phoneme sequences are generated from orthographic transcripts using a multilingual grapheme-to-phoneme (G2P) pipeline. We use G2P+ \cite{goriely2025ipa} to generate IPA-based phonemic transcriptions across languages. Although generated phonemic references introduce limitations due to potentially incorrect, oversimplified phoneme mappings, incomplete phonetic detail, and inconsistencies across languages, they provide a scalable framework for multilingual evaluation. These generated phoneme sequences serve as the ground-truth references for all experiments.

\paragraph{Soft PER} We evaluate phoneme recognition accuracy using phoneme error rate (PER). However, because phoneme realizations can vary across accents, dialects, and languages without reflecting true recognition failures, we also introduce and evaluate a Soft PER metric that reduces penalties for linguistically similar substitutions. This evaluation uses a two-tier phoneme mapping to (1) establish broad interchangeable phonemic variation and (2) create language-specific pronunciation overlaps that are not global equivalents.  A single equivalence tier is insufficient because many phonetic relationships are not identical between languages.

The first tier groups phonemes into transitive equivalence classes using two allophonic resources. AlloVera \cite{mortensen2020allovera} provides language-specific mappings between surface phones and canonical phonemes across 14 languages (e.g., aspirated /\textipa{p\super h}/$\rightarrow$/\textipa{p}/ where a strongly breathed ``p'' sound is treated as a standard /\textipa{p}/, devoiced /\textipa{b}/\textsubring{}$\rightarrow$/\textipa{b}/ where a partially unvoiced ``b'' is mapped to its canonical form, etc.), with English given priority, because English defines the largest proportion of the demographic evaluation data, when a surface phone appears in multiple languages. We then use articulatory feature representations from PHOIBLE \cite{phoible} to identify pairs of canonical phonemes that differ in only one of 34 articulatory features. These pairs are merged into shared equivalence classes. 
Through this grouping, phonemes may belong to the same class even if they differ in more than one feature, as long as they are connected through phonemes that differ by only one feature at each step. Together, these mappings create 62 equivalence classes spanning 254 phones. Table \ref{appendix:tier_one} in Appendix~\ref{appendix:soft-per} provides the full mapping for the first tier.


The second tier captures non-transitive cross-language similarities, derived from AlloVera \cite{mortensen2020allovera}. When the same surface phone maps to different canonical phonemes across languages (e.g., the flap sound /\textipa{R}/ is mapped to /\textipa{t}/ in English words such as ``water,'' but to /\textipa{r}/ in Spanish words such as ``pero''), those canonical phonemes are recorded as a direct pair without being merged into a shared equivalence class. This preserves the individual relations /\textipa{t}/$\approx$/\textipa{R}/ and /\textipa{R}/$\approx$/\textipa{r}/ without implying /\textipa{t}/$\approx$/\textipa{r}/ and keeping language-specific relationships. A total of 90 such language-specific pairs are extracted from cross-language allophonic overlaps. These pairs are scored as equivalent within their respective language mappings, but are not transitively combined. Table \ref{appendix:tier_two} in Appendix \ref{appendix:soft-per} provides the full mapping for the second tier.

For example, Tier 1 places devoiced /\textipa{b}/\textsubring{} and canonical /\textipa{b}/ in the same equivalence class, so substituting one for the other incurs no error. In contrast, /\textipa{d}/ and /\textipa{t}/ belong to different Tier 1 classes and would initially be counted as a substitution error. Tier 2 contains this pair as a language-specific similarity in English, Kazakh, Russian, and Vietnamese, so the substitution isn't counted when it happens in one of these languages. Unlike Tier 1 equivalences, this relation is not extended transitively to other phonemes.

During evaluation, Tier 1 equivalence classes are applied universally across all languages, while Tier 2 similarities are only applied when relevant to the target language. Substitutions within either similarity tier receive a zero penalty when computing Soft PER, allowing the metric to better distinguish genuine recognition errors from acceptable phonetic variation.

\section{Results}
\subsection{Language Performance Evaluation}

\begin{figure*}[!ht]
    \centering
    \begin{subfigure}{0.48\textwidth}
        \centering
        \includegraphics[width=\linewidth]{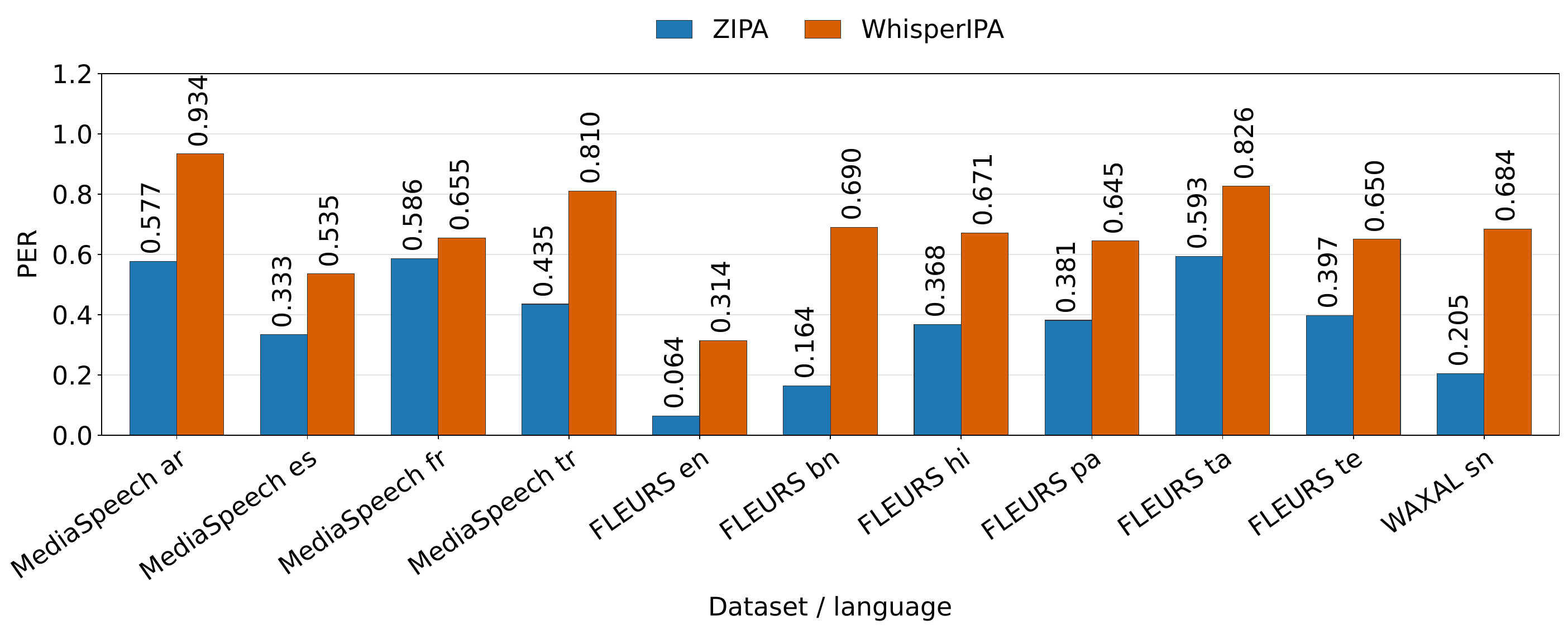}
        \caption{Standard PER across languages for ZIPA and WhisperIPA.}
        \label{fig:language_exact_per}
    \end{subfigure}
    \hfill
    \begin{subfigure}{0.48\textwidth}
        \centering
        \includegraphics[width=\linewidth]{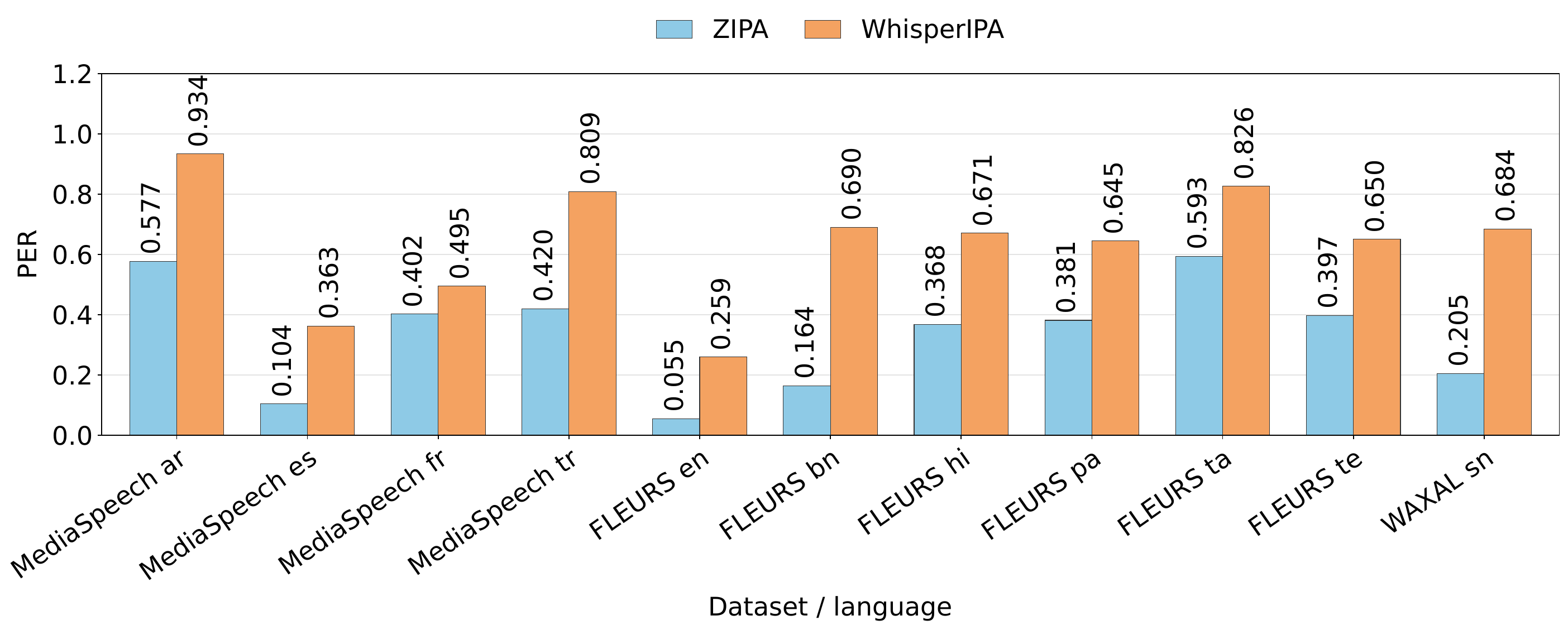}
        \caption{Soft PER across languages for ZIPA and WhisperIPA.}
        \label{fig:language_soft_per}
    \end{subfigure}

    \caption{Comparison of standard PER and Soft PER across evaluated languages: (ar=Arabic, es=Spanish, fr=French, tr=Turkish, en=English, bn=Bangla, hi=Hindi, pa=Panjabi, ta=Tamil, te=Telugu, and sn=Shona).}
    \label{fig:language_per_comparison}
\end{figure*}

\begin{figure*}[!ht]
    \centering
    \begin{subfigure}{0.48\textwidth}
        \centering
        \includegraphics[width=\linewidth]{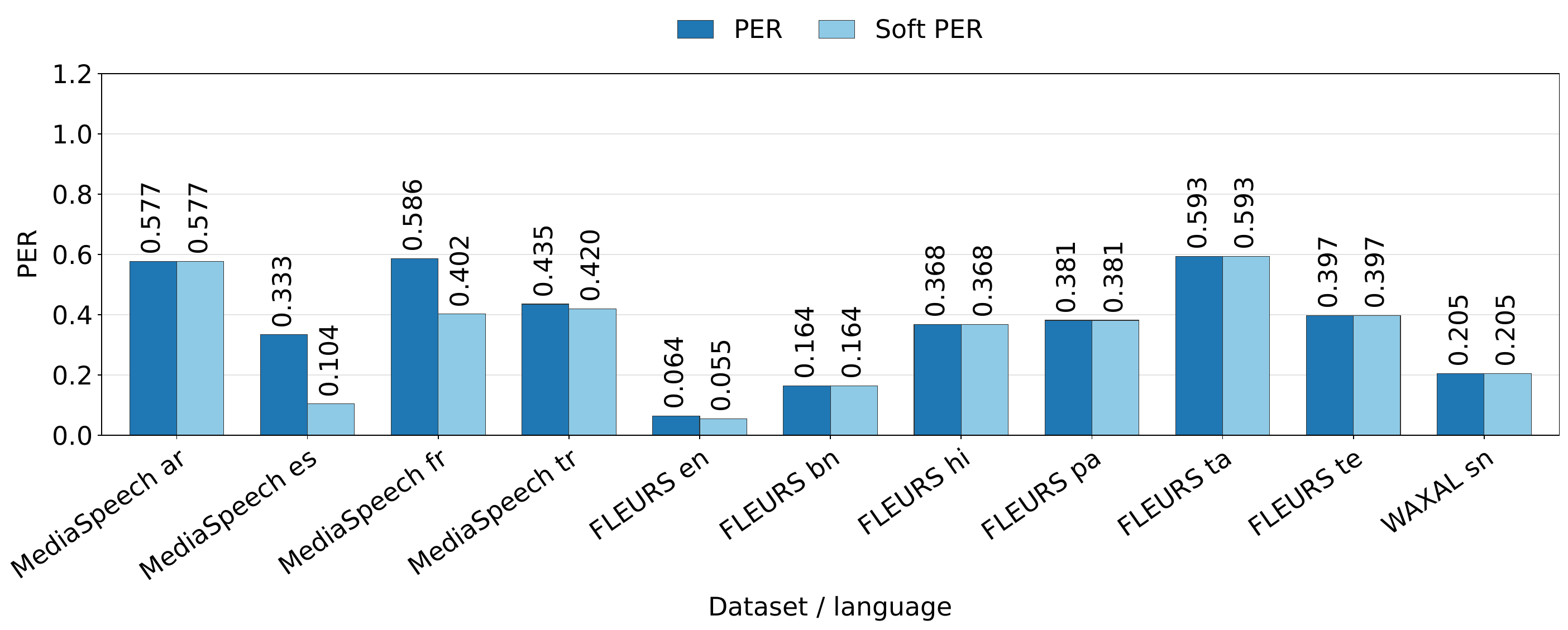}
        \caption{Standard PER and Soft PER comparison for ZIPA.}
        \label{fig:zipa_exact_soft}
    \end{subfigure}
    \hfill
    \begin{subfigure}{0.48\textwidth}
        \centering
        \includegraphics[width=\linewidth]{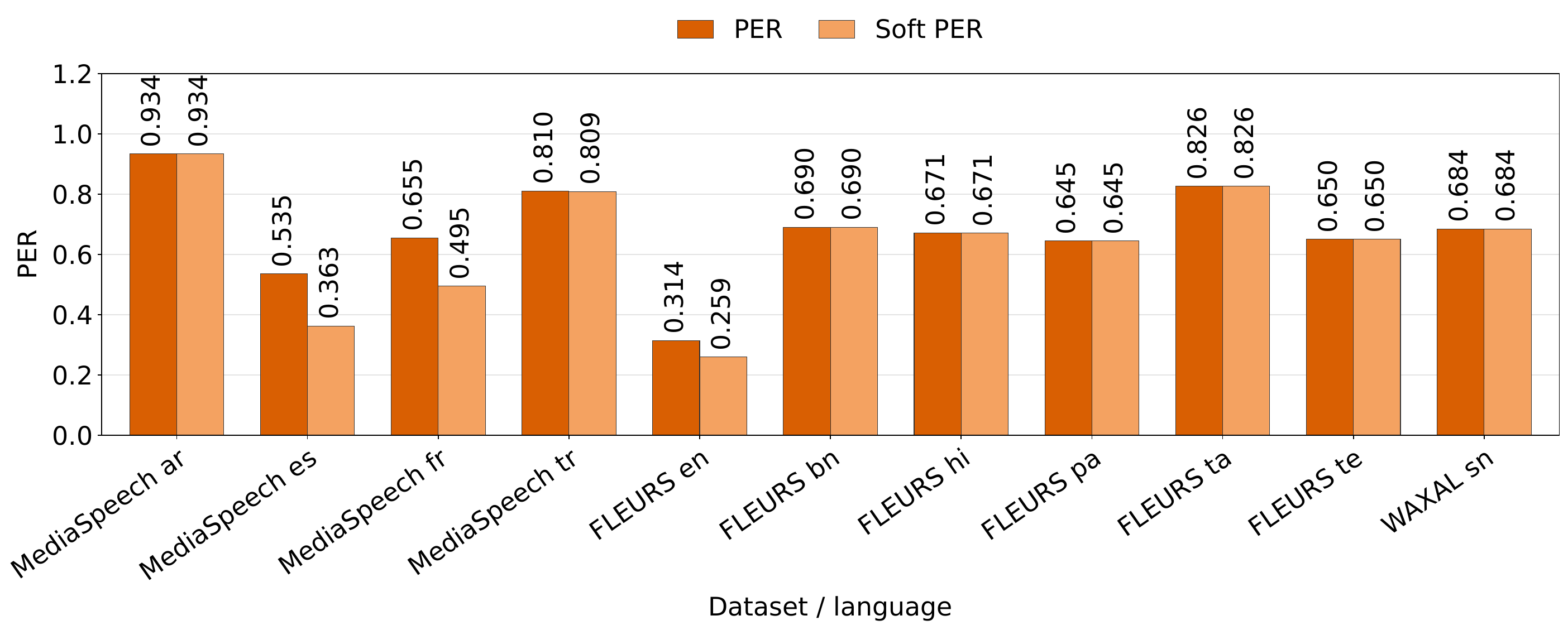}
        \caption{Standard PER and Soft PER comparison for WhisperIPA.}
        \label{fig:whisper_exact_soft}
    \end{subfigure}

    \caption{Effect of Soft PER on language-level evaluation for each model.}
    \label{fig:language_soft_reduction}
\end{figure*}

We first evaluate model performance across languages to understand how IPA-based ASR system accuracy varies in a multilingual setting. This analysis compares standard PER and Soft PER for both ZIPA and WhisperIPA across the available language datasets. Standard PER measures exact phoneme-level mismatch, while Soft PER applies the equivalence mappings described in Section~\ref{sec:exp-setup} to reduce penalties for linguistically similar phoneme substitutions.

Figure~\ref{fig:language_per_comparison} compares ZIPA and WhisperIPA across the evaluated languages using both standard PER and Soft PER. ZIPA achieves lower error rates than WhisperIPA for every language under both metrics, indicating stronger overall phoneme-recognition performance in this multilingual evaluation. Under standard PER, English has the lowest error rate for both models. For ZIPA, the highest error rates occur for Tamil, French, and Arabic, whereas WhisperIPA produces its highest error rates for Arabic, Tamil, and Turkish. These results demonstrate substantial variation in performance across languages.

Under Soft PER, ZIPA continues to outperform WhisperIPA for every language, although the effect of the soft-equivalence mapping varies considerably across languages. The largest reductions occur for Spanish and French. For ZIPA, Spanish decreases from 0.333 to 0.104 and French decreases from 0.586 to 0.402. WhisperIPA shows similar reductions, with Spanish decreasing from 0.535 to 0.363 and French decreasing from 0.655 to 0.495. English also improves under Soft PER, particularly for WhisperIPA, while Turkish shows only a small reduction for ZIPA and a negligible change for WhisperIPA. In contrast, the values for Arabic, Bangla, Hindi, Panjabi, Tamil, Telugu, and Shona remain unchanged. Thus, Soft PER affects only a subset of the evaluated languages rather than reducing errors uniformly across the multilingual evaluation.

Figure~\ref{fig:language_soft_reduction} presents the effect of Soft PER separately for each model. For ZIPA, the reduction is largest for Spanish and French, smaller for Turkish and English, and absent for the remaining languages. WhisperIPA follows a similar pattern, with substantial reductions for Spanish and French, a smaller reduction for English, and almost no change for Turkish. The other languages remain unchanged. Despite these language-specific reductions, WhisperIPA continues to produce higher error rates than ZIPA across all languages.


\paragraph{Key Takeaways:}
Overall, ZIPA consistently outperforms WhisperIPA across all evaluated languages. ZIPA achieves an average standard PER of 0.373, compared with 0.674 for WhisperIPA, corresponding to an absolute difference of 0.301. Under Soft PER, the averages decrease to 0.333 and 0.639, respectively, while the absolute difference remains nearly unchanged at 0.305. Thus, Soft PER lowers the aggregate error rates but does not narrow the performance gap between the models. This suggests that ZIPA's advantage is not explained primarily by the phonemically similar substitutions captured by the current Soft PER mapping.

Considering English, Spanish, and French as high-resource languages and Hindi, Bangla, Panjabi, Tamil, Telugu, Arabic, Turkish, and Shona as low-resource languages, average error rates are higher for the low-resource group. Averaged across both models, standard PER is 0.565 for low-resource languages and 0.415 for high-resource languages, representing a 36.2\% relative increase. Under Soft PER, the corresponding averages are 0.563 and 0.280, producing an absolute difference of 0.284 and a relative difference of approximately 101\%.

The widening relative gap under Soft PER is driven primarily by the substantial reductions observed for Spanish and French, both of which belong to the high-resource group. English also improves, although by a smaller amount. In comparison, most low-resource languages remain unchanged, with only a small reduction for Turkish. One possible explanation is that the AlloVera-derived equivalence mappings, which prioritize English when mappings overlap across languages, do not adequately represent phonemic variation in many of the low-resource languages. Differences in model architecture, training objectives, and multilingual training data may also contribute to the observed pattern. Overall, Soft PER identifies phonemically similar substitutions in a limited subset of languages but does not change the model ranking or reduce the average performance disparity between the high- and low-resource groups.

\subsection{Demographic Performance Evaluation}
\label{sec:demographic-performance}
We next analyze how phoneme recognition performance varies across demographic groups using CORAAL, EdAcc, SVC, and WAXAL, focusing on gender, age, ethnicity, and accent/dialect. We interpret demographic patterns within each dataset rather than across datasets, since dataset characteristics may contribute more to performance variation than the demographic attributes themselves. We report bootstrap 95\% confidence intervals for all group means shown as error bars in the figures and test every pairwise group contrast with a Welch's $t$-test. Full pairwise results are reported by demographic attribute in Tables~\ref{tab:sig-gender-all}--\ref{tab:sig-accent-svc} in Appendix~\ref{appendix:welch-ttest}.

\paragraph{Gender:}


\begin{figure*}[!ht]
    \centering

    \begin{subfigure}{0.48\textwidth}
        \centering
        \includegraphics[width=\linewidth]{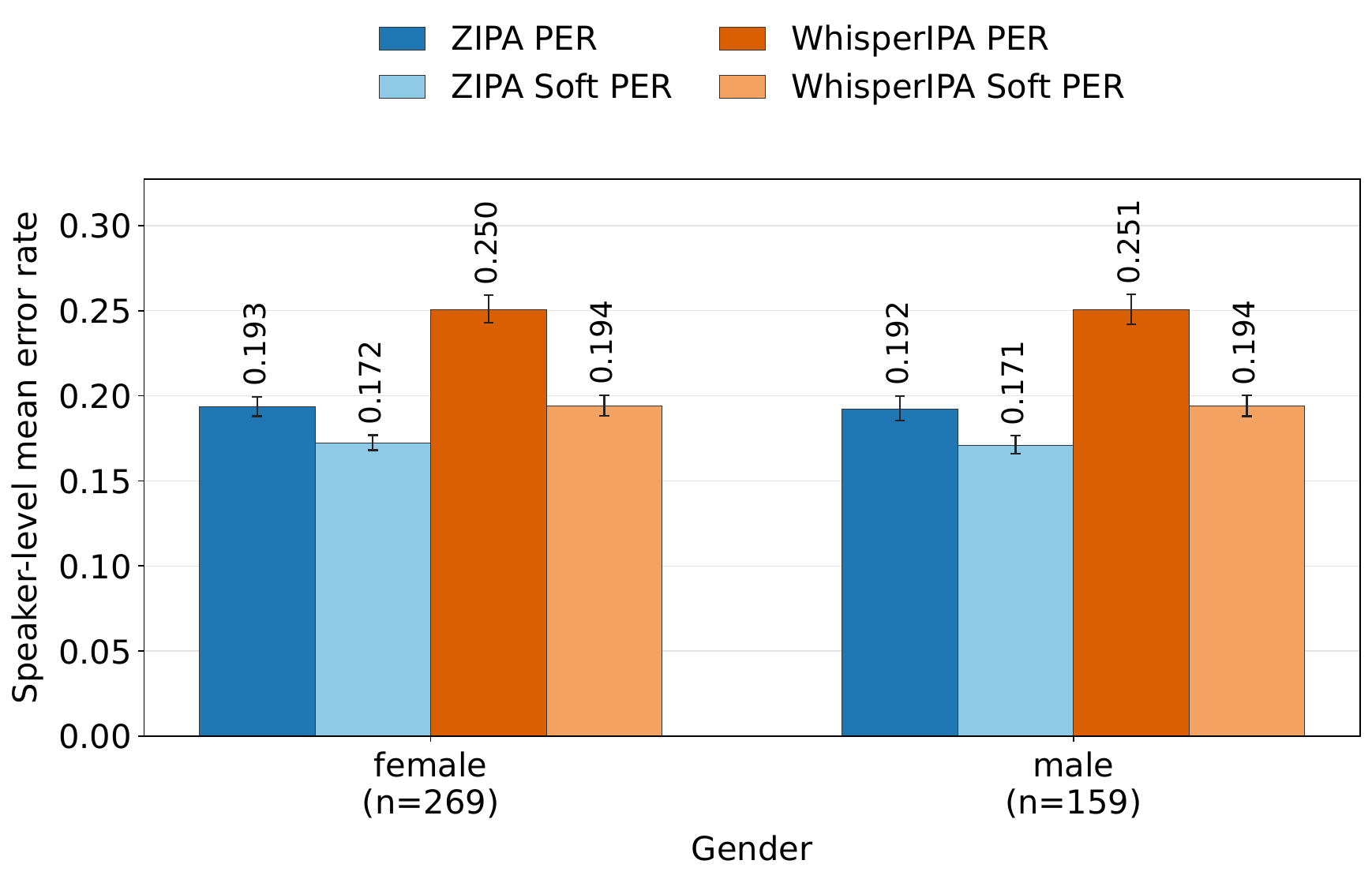}
        \caption{PER and Soft PER by gender for SVC Dataset}
        \label{fig:svc_gender_per}
    \end{subfigure}
    \hfill
    \begin{subfigure}{0.48\textwidth}
        \centering
        \includegraphics[width=\linewidth]{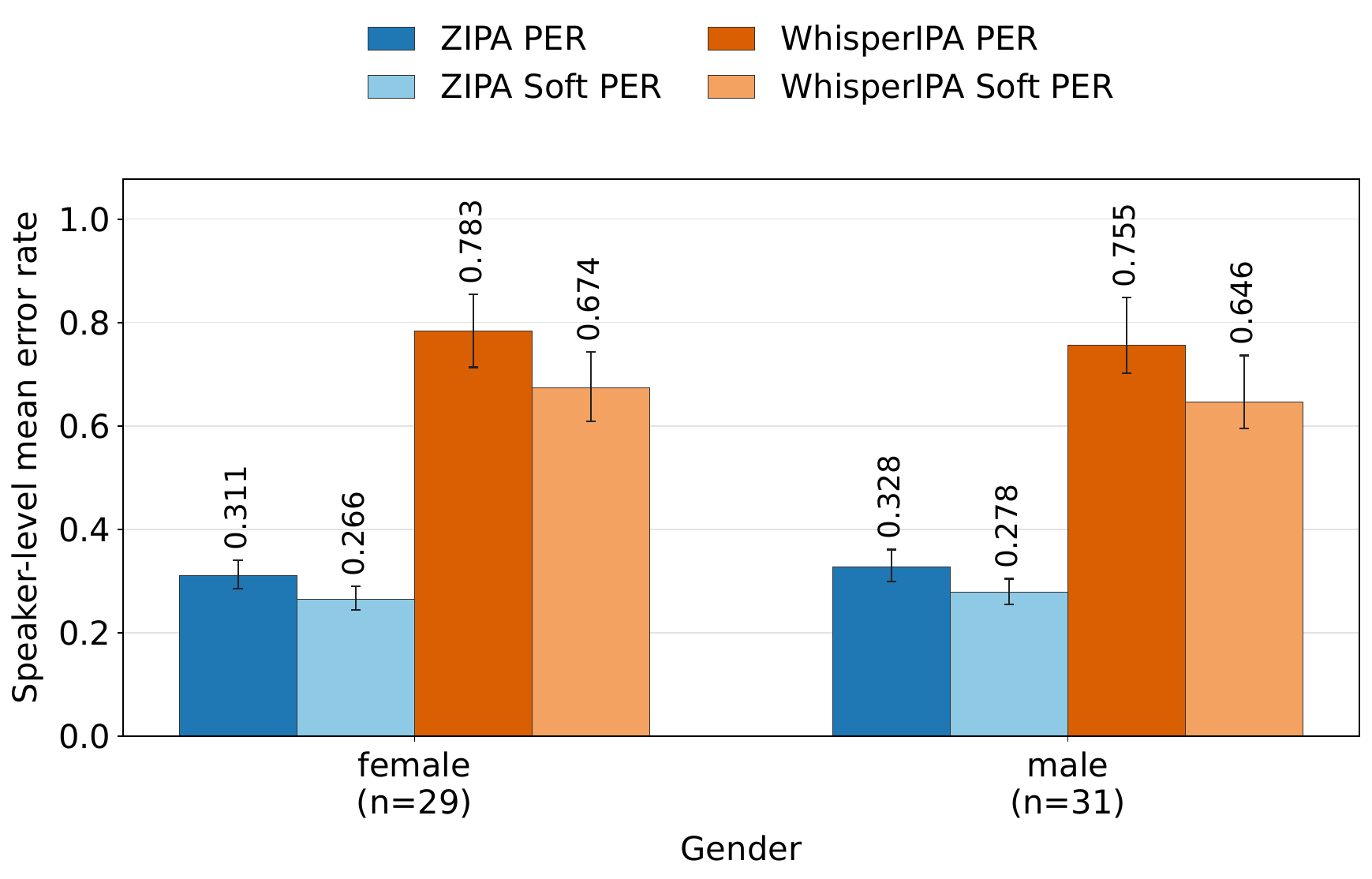}
        \caption{PER and Soft PER by gender for EdAcc Dataset}
        \label{fig:EdAcc_gender_per}
    \end{subfigure}

    \par\medskip

    \begin{subfigure}{0.48\textwidth}
        \centering
        \includegraphics[width=\linewidth]{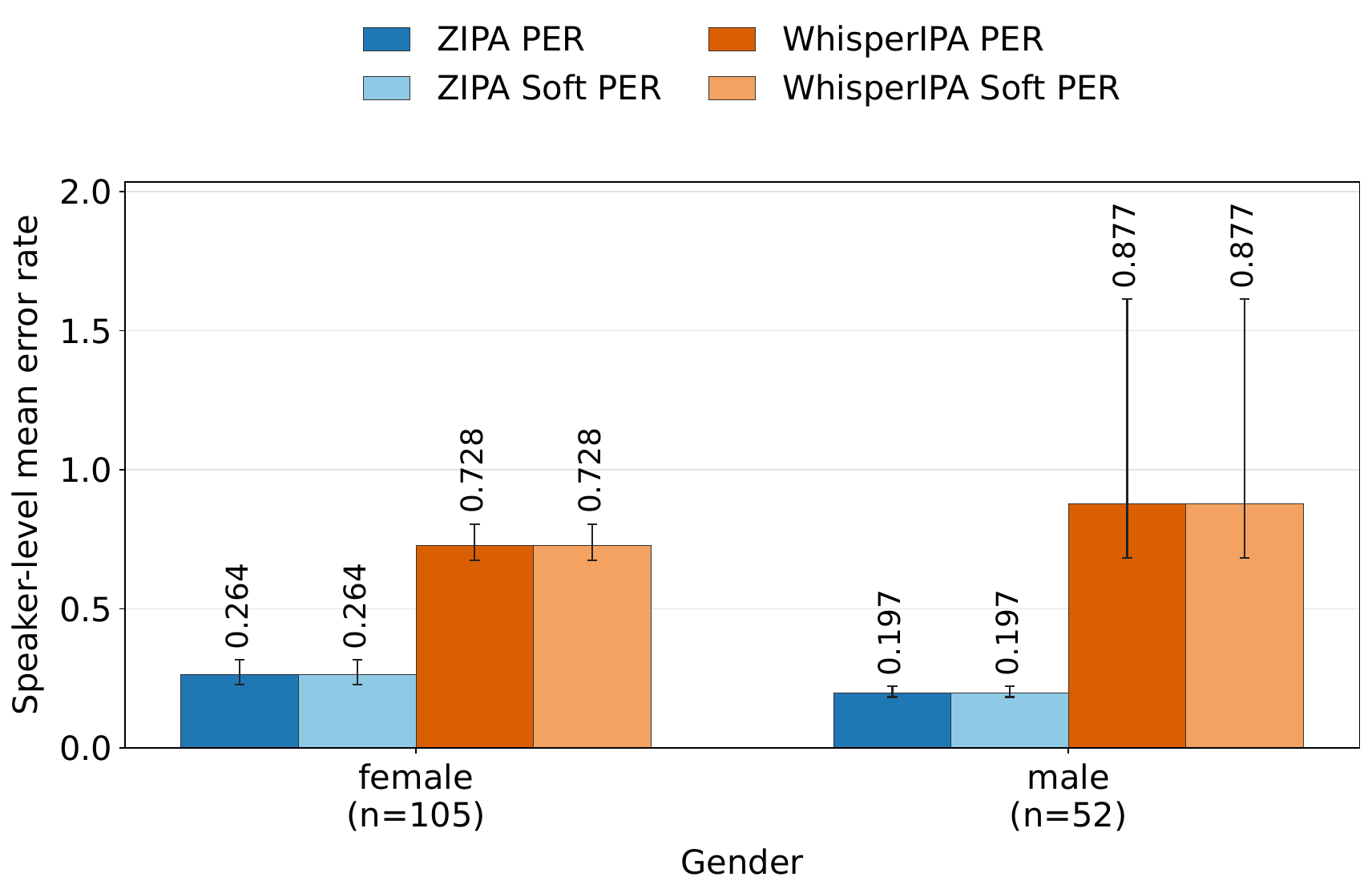}
        \caption{PER and Soft PER by gender for WAXAL Dataset}
        \label{fig:waxal_gender_per}
    \end{subfigure}
    \hfill
    \begin{subfigure}{0.48\textwidth}
        \centering
        \includegraphics[width=\linewidth]{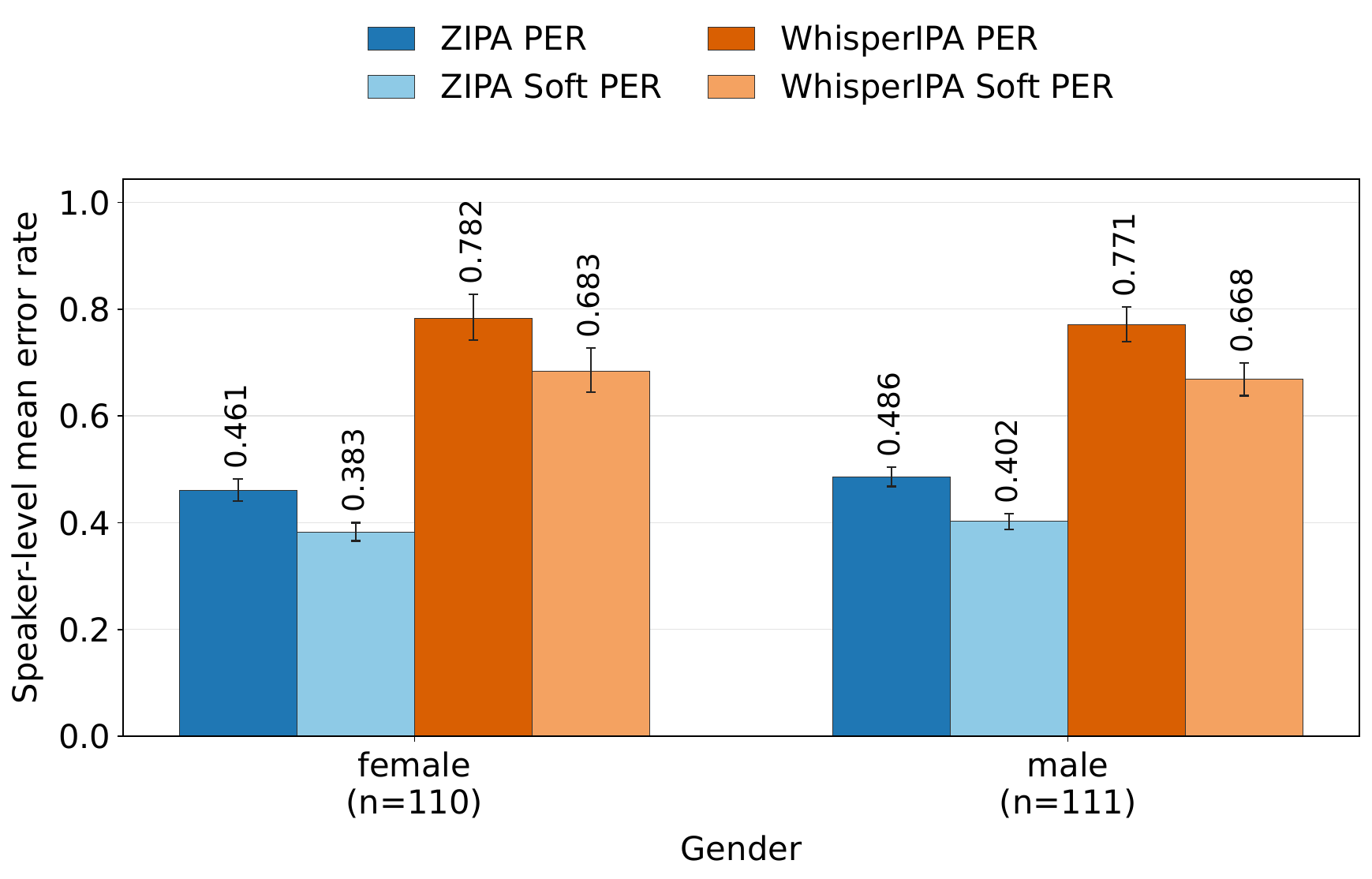}
        \caption{PER and Soft PER by gender for CORAAL Dataset}
        \label{fig:coraal_gender_per}
    \end{subfigure}

    \caption{PER and Soft PER comparison by gender across demographic datasets. Bars and numerical labels represent speaker-level mean error rates. Error bars indicate 95\% BCa bootstrap confidence intervals, with percentile bootstrap intervals used when BCa estimation does not converge or when $n<10$.}
    \label{fig:gender_per}
\end{figure*}

Figure~\ref{fig:gender_per} presents PER and Soft PER across gender groups in SVC, EdAcc, WAXAL, and CORAAL datasets. In SVC, EdAcc, and CORAAL datasets, male and female speakers exhibit similar error rates, with no consistent direction of disparity across models or metrics. WAXAL shows the only significant gender contrast: PER and Soft PER for ZIPA are significantly higher for female speakers ($+0.067$, $p<0.01$; Table~\ref{tab:sig-gender-all}). In contrast, WhisperIPA shows a larger but not significant descriptive gap in the opposite direction ($-0.149$), likely reflecting the small male sample ($n=52$). Soft PER lowers absolute error rates while preserving within-dataset patterns (identical to PER for WAXAL, since none of its substitutions fall under the soft-equivalence mapping). Overall, aside from this single ZIPA effect in WAXAL, the results show limited evidence of systematic gender-based disparity.

\paragraph{Age:}
\begin{figure}[!ht]
    \centering

    \begin{subfigure}{\columnwidth}
        \centering
        \includegraphics[width=\linewidth]{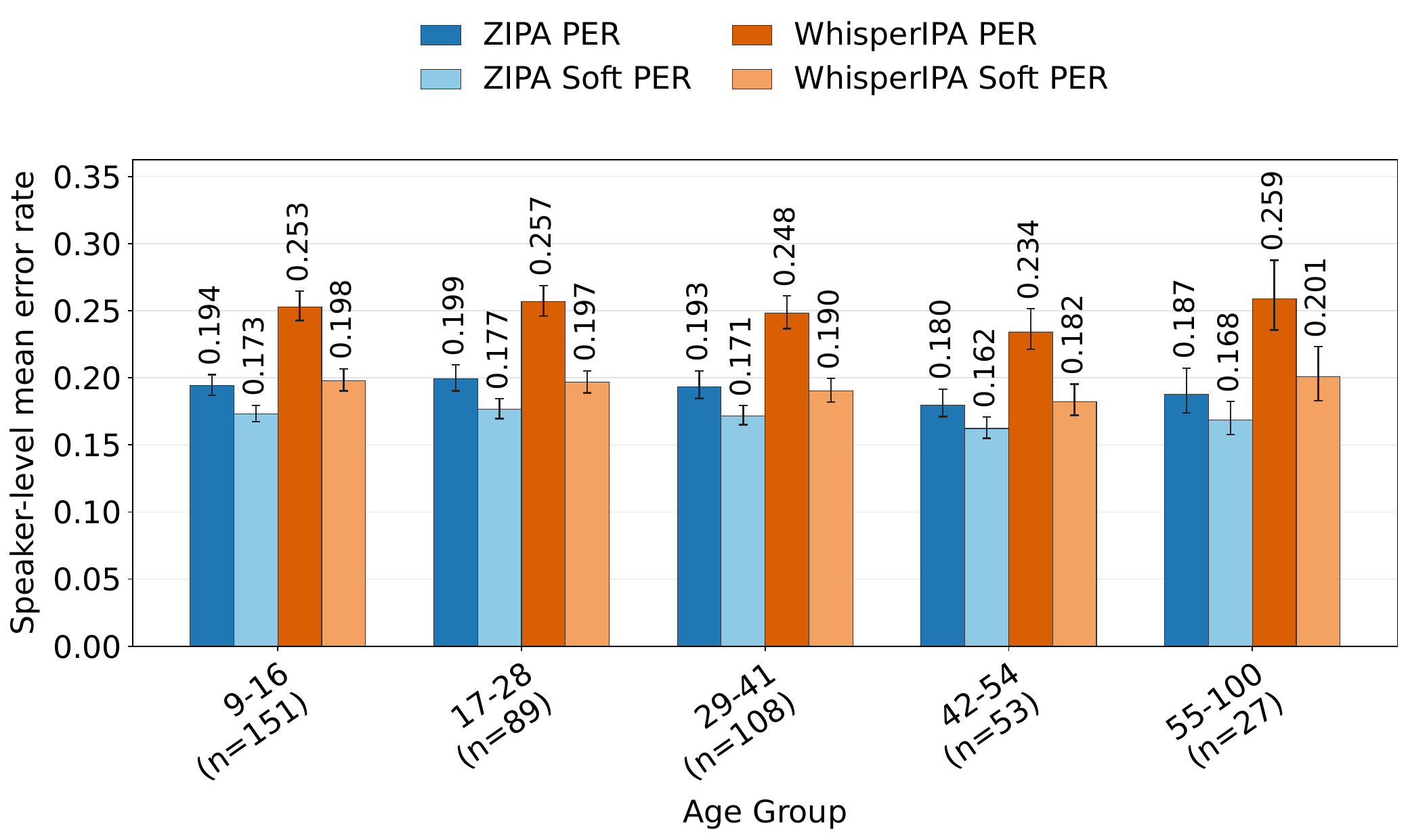}
        \caption{PER and Soft PER by age group for the SVC dataset.}
        \label{fig:svc_age_per}
    \end{subfigure}
    \hfill
    \begin{subfigure}{\columnwidth}
        \centering
        \includegraphics[width=\linewidth]{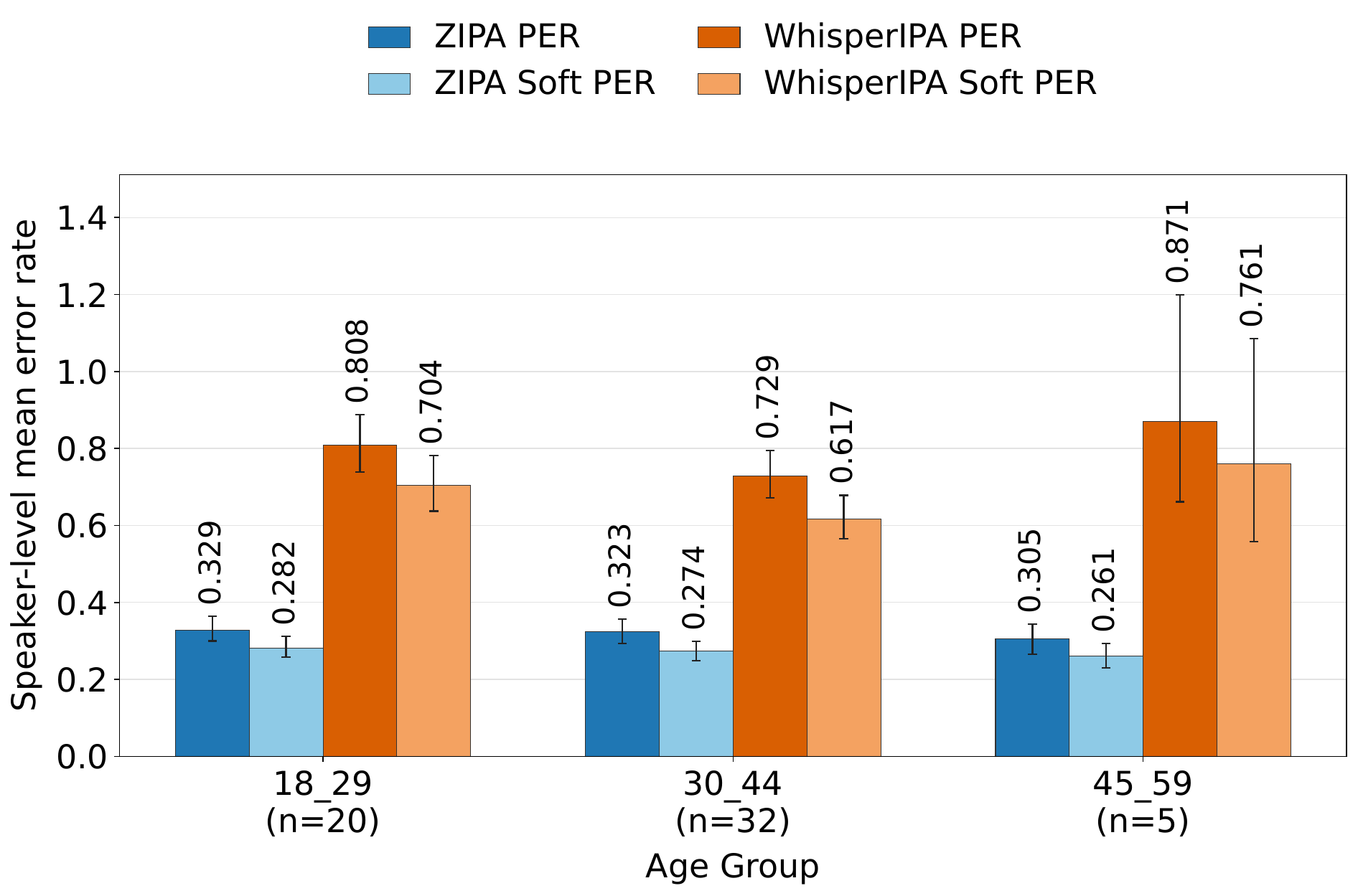}
        \caption{PER and Soft PER by age group for the EdAcc dataset.}
        \label{fig:EdAcc_age_per}
    \end{subfigure}
    \hfill
    \begin{subfigure}{\columnwidth}
        \centering
        \includegraphics[width=\linewidth]{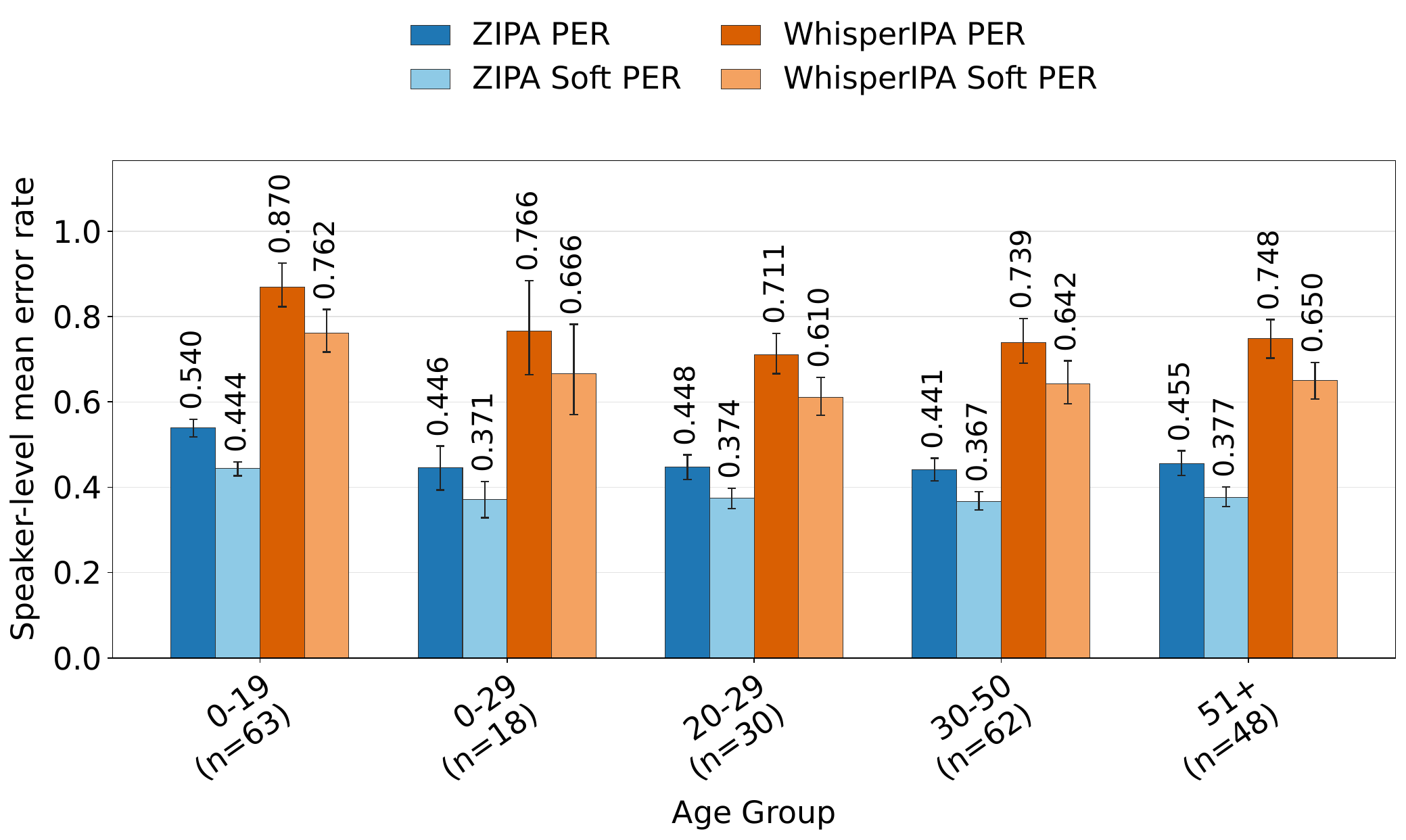}
        \caption{PER and Soft PER by age group for the CORAAL dataset.}
        \label{fig:coraal_age_per}
    \end{subfigure}

    \caption{PER and Soft PER comparison by age group across datasets. Bars and numerical labels represent speaker-level mean error rates. Error bars indicate 95\% BCa bootstrap confidence intervals, with percentile bootstrap intervals used when BCa estimation does not converge or when $n<10$.}
    \label{fig:age_per}
\end{figure}

Figure~\ref{fig:age_per} presents PER and Soft PER across age groups in SVC, EdAcc, and CORAAL datasets. In SVC, both models show small absolute differences across age groups (Figure~\ref{fig:svc_age_per}), though several are statistically significant: the 42--54 group has significantly lower PER than the 17--28 group (all four metrics, $p<0.05$, $p<0.01$ for ZIPA) and the 9--16 group (ZIPA PER/Soft PER and WhisperIPA Soft PER, $p<0.05$; Table~\ref{tab:sig-age-svc}), with no other SVC pair reaching significance. In EdAcc, ZIPA is stable across ages, while WhisperIPA peaks for the 45--59 group (Figure~\ref{fig:EdAcc_age_per}); this should be read cautiously, as the group has only five speakers and none of EdAcc's three pairwise age contrasts reach significance (Table~\ref{tab:sig-age-edacc}). CORAAL\footnote{CORAAL component corpora use nonuniform age bins; although the speaker groups are disjoint, the source label 0–29 overlaps in range with 0–19 and 20–29 and may partly capture subcorpus differences. See the \href{https://talkbank.org/ca/access/0docs/CORAAL.pdf}{CORAAL User Guide}.}
 shows the clearest pattern: the 0--19 group has significantly higher PER than the 20--29, 30--50, and 51+ groups on all four metrics ($p<0.01$, mostly $p<0.001$) and higher ZIPA PER/Soft PER than the 0--29 group ($p<0.01$), though the WhisperIPA contrast with 0--29 is not significant ($p=0.11$, $p=0.13$; Table~\ref{tab:sig-age-coraal}). No other pair differs significantly. Soft PER lowers absolute error rates while largely preserving these patterns. Overall, age effects are dataset-dependent: a robust effect for the youngest speakers in CORAAL, a modest but significant effect in SVC's 42--54 group, and no significant effect in EdAcc despite a large descriptive gap.

\paragraph{Ethnicity:}

\begin{figure*}[!ht]
    \centering
    \begin{subfigure}{0.48\textwidth}
        \centering
        \includegraphics[width=\linewidth]{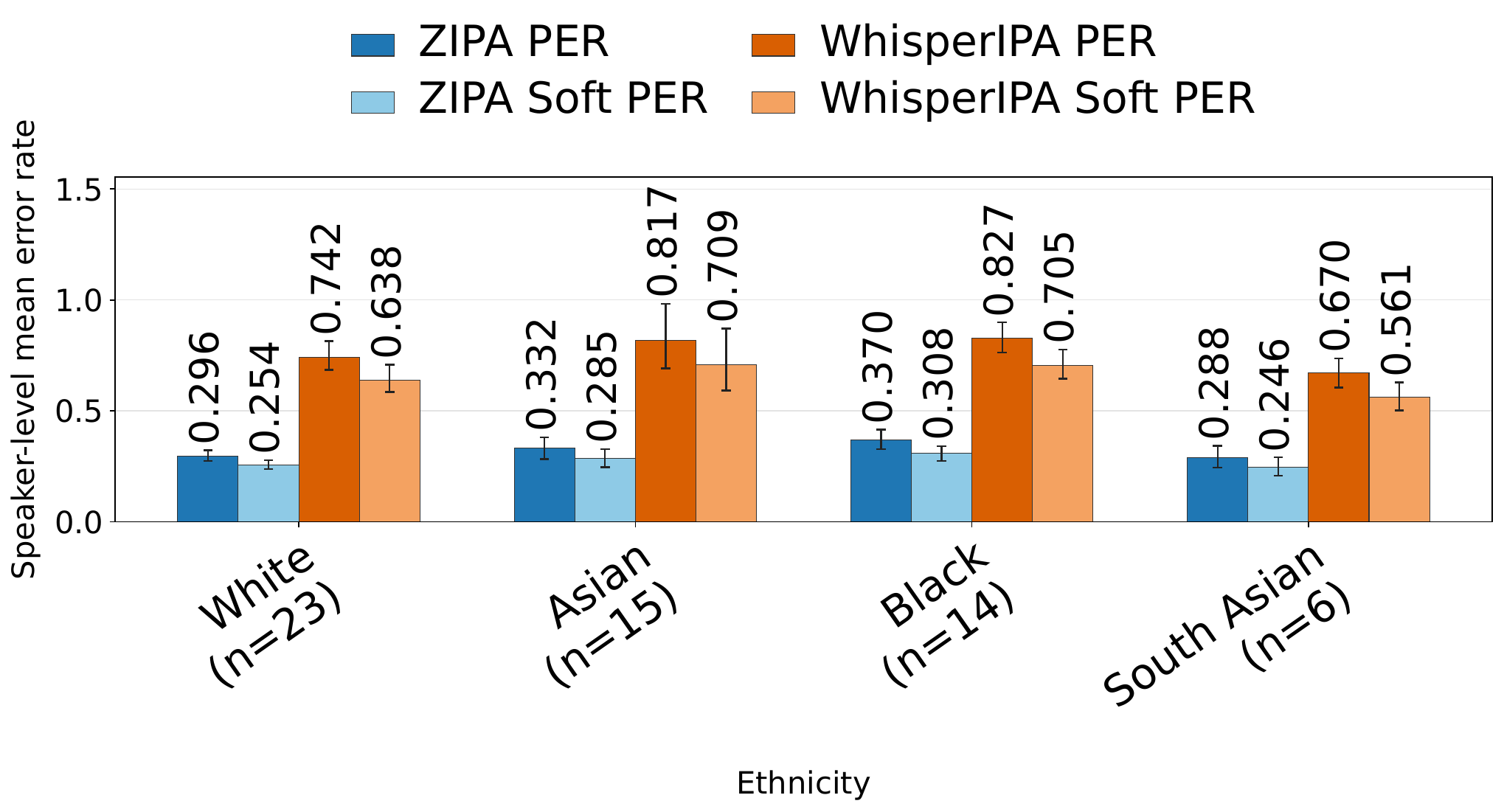}
        \caption{PER and Soft PER by ethnicity group for the EdAcc dataset.}
        \label{fig:EdAcc_ethnicity_per}
    \end{subfigure}
    \hfill
    \begin{subfigure}{0.48\textwidth}
        \centering
        \includegraphics[width=\linewidth]{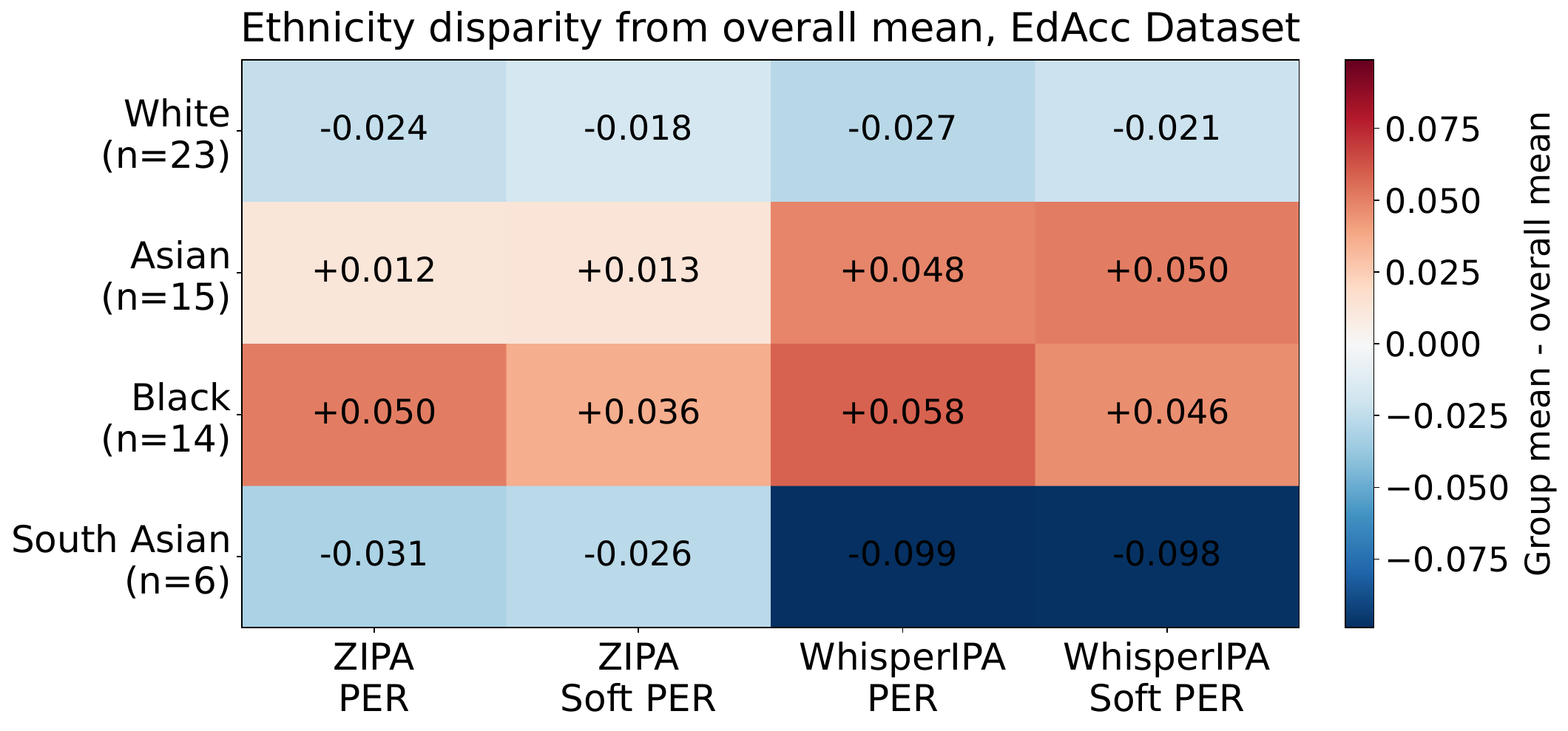}
        \caption{Ethnicity disparity from the overall mean for the EdAcc dataset.}
        \label{fig:EdAcc_ethnicity_disparity}
    \end{subfigure}

    \caption{PER, Soft PER, and group-level disparity by ethnicity for the EdAcc dataset. Bars and numerical labels represent speaker-level mean error rates. Error bars indicate 95\% BCa bootstrap confidence intervals, with percentile bootstrap intervals used when BCa estimation does not converge or when $n<10$.}
    \label{fig:ethnicity_per}
\end{figure*}

Figure~\ref{fig:ethnicity_per} presents PER and Soft PER across ethnicity groups in the EdAcc dataset. ``White'', ``Asian'', ``Black'', and ``South Asian'' are possible self-selected ethnicity descriptors in EdAcc.\footnote{\url{https://groups.inf.ed.ac.uk/EdAcc/EdAcc_statement.pdf}} As shown in Figure~\ref{fig:EdAcc_ethnicity_per}, WhisperIPA produces substantially higher error rates than ZIPA across all groups, consistent with the model-level differences observed throughout the demographic evaluation.

Pairwise Welch's $t$-tests (Table~\ref{tab:sig-ethnicity-edacc}) show ``Black'' speakers have significantly higher PER than ``South Asian'' speakers for ZIPA PER, WhisperIPA PER ($+0.157$, $p<0.01$), and WhisperIPA Soft PER ($+0.144$, $p<0.05$), and higher PER than ``White'' speakers for ZIPA PER and Soft PER ($+0.074$, $+0.054$; $p<0.05$). No other contrast is significant: ``Asian'' speakers show an elevated disparity from the overall mean (Figure~\ref{fig:EdAcc_ethnicity_disparity}) but are not significantly different from any individual group, including ``Black'' speakers. All estimates rest on small samples ($n=6$--$23$) and should be interpreted accordingly.



\paragraph{Accent/Dialect:}

\begin{figure*}[!ht]
    \centering
    \begin{subfigure}{0.48\textwidth}
        \centering
        \includegraphics[width=\linewidth]{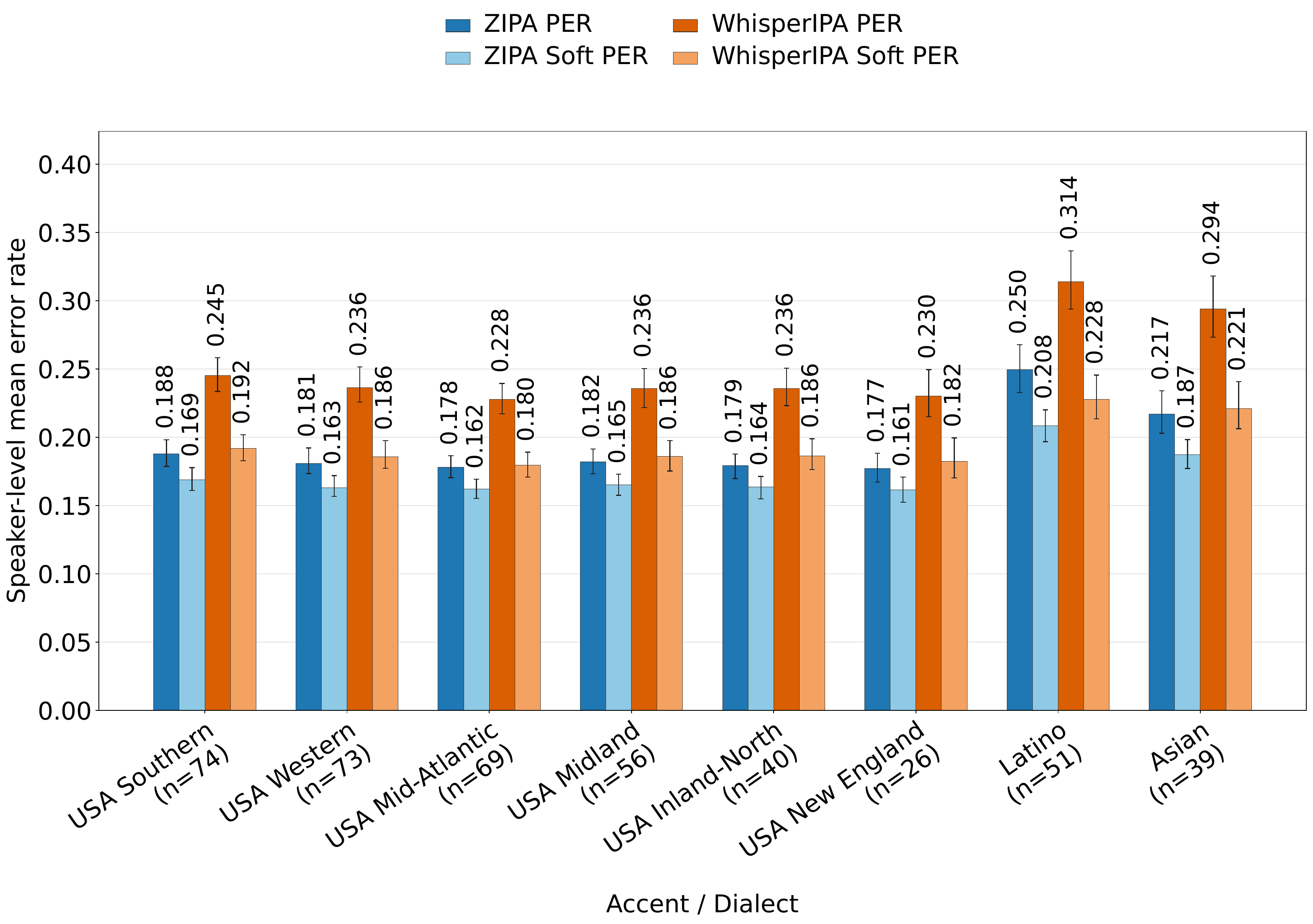}
        \caption{PER and Soft PER by accent/dialect group for the SVC dataset.}
        \label{fig:svc_accent_per}
    \end{subfigure}
    \hfill
    \begin{subfigure}{0.48\textwidth}
        \centering
        \includegraphics[width=\linewidth]{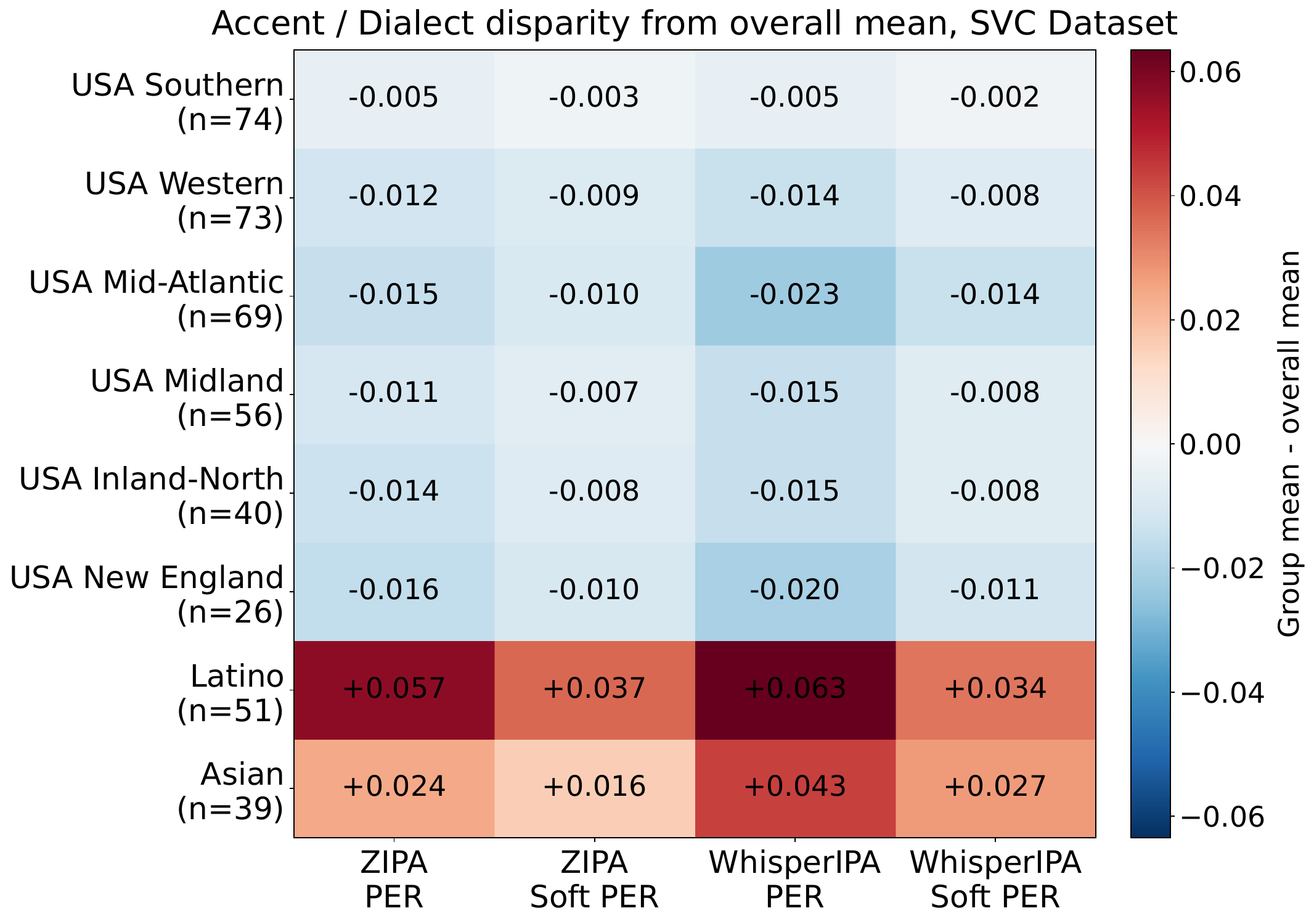}
        \caption{Accent/dialect disparity from the overall mean for the SVC dataset.}
        \label{fig:svc_accent_disparity}
    \end{subfigure}

    \caption{PER, Soft PER, and group-level disparity by accent/dialect for the SVC dataset. Bars and numerical labels represent speaker-level mean error rates. Error bars indicate 95\% BCa bootstrap confidence intervals, with percentile bootstrap intervals used when BCa estimation does not converge or when $n<10$.}
    \label{fig:accent_per}
\end{figure*}

Figure~\ref{fig:accent_per} presents PER and Soft PER across accent and dialect groups in the SVC dataset. In SVC, six regional groups---``USA Southern'', ``USA Western'', ``USA Mid-Atlantic'', ``USA Midland'', ``USA Inland-North'', and ``USA New England''---represent native speakers of American English of any racial or ethnic group. The ``Asian'' and ``Latino'' groups represent non-native speakers from Asia and Latin America who reside in any region of the USA.

As shown in Figure~\ref{fig:svc_accent_per}, most USA regional groups show similar or slightly below-average error for both models, while the ``Latino'' and ``Asian'' groups have the highest error, particularly for WhisperIPA --- reflecting greater difficulty with accents that differ from SVC's native regional varieties. The disparity heatmap (Figure~\ref{fig:svc_accent_disparity}) and pairwise Welch's $t$-tests confirm this is a robust effect (Table~\ref{tab:sig-accent-svc}): ``Latino'' speakers have the largest positive disparity from the overall mean ($+0.057$/$+0.037$ ZIPA PER/Soft PER, $+0.063$/$+0.034$ WhisperIPA), and both ``Latino'' and ``Asian'' speakers have significantly higher PER than nearly every individual USA regional group ($p<0.001$ in most cases). ``Latino'' speakers are also significantly higher than ``Asian'' speakers under ZIPA ($+0.033$, $+0.021$; $p<0.05$), though not under WhisperIPA. The six USA regional groups are largely indistinguishable from one another, with only a handful of pairwise contrasts reaching significance. Soft PER reduces the magnitude of these disparities but preserves the relative pattern, indicating the differences are not fully explained by the minor phonemic variation captured by the soft-equivalence mapping.

\paragraph{Key Takeaways:}
Overall, gender shows limited evidence of disparity, aside from a small but significant ZIPA effect in WAXAL. Accent and ethnicity show the most robust effects: in SVC, ``Latino'' and ``Asian'' speakers have significantly higher PER than nearly every USA regional group across both models; in EdAcc, ``Black'' speakers have significantly higher PER than ``South Asian'' speakers (and, for ZIPA, ``White'' speakers as well), while ``Asian'' speakers are not significantly different from any other group despite an elevated disparity from the overall mean. Age effects are more dataset-dependent: a robust elevation for the youngest speakers in CORAAL, a modest but significant effect in SVC's 42--54 group, and no significant effect in EdAcc despite a large descriptive gap. Several estimates rest on small samples and should be interpreted cautiously. Overall, demographic performance patterns reflect an interaction of model behavior, dataset characteristics, and group representation rather than any single attribute.

\paragraph{PER vs.\ Soft PER:}
Soft PER generally reduces absolute error rates while preserving the relative performance patterns observed across demographic groups. If the disparities measured by standard PER were driven primarily by phonologically similar substitutions, including variations associated with accents or dialects, applying the soft-equivalence mapping would be expected to narrow the gaps between groups substantially. Instead, the persistence of similar group-level patterns under both metrics suggests that the observed differences also involve recognition errors not covered by the current mapping. The limited effect of Soft PER may additionally reflect incomplete coverage of demographic and cross-lingual phonemic variation within the equivalence classes. In this evaluation, standard PER and Soft PER therefore lead to broadly similar qualitative conclusions about relative group performance. Standard PER remains useful for comparison with prior work, whereas Soft PER provides a complementary measure for examining how acceptable phonemic variation contributes to the reported error rates.

\section{Conclusion}

In this work, we evaluate bias and performance variation in IPA-based phoneme recognition models across languages and speaker demographics. We compare two open-source IPA transcription systems, ZIPA and WhisperIPA, using both standard PER and a proposed Soft PER metric that reduces penalties for linguistically similar phoneme substitutions. Across the multilingual evaluation, ZIPA consistently achieves lower error rates than WhisperIPA, while both models show substantial variation across languages. This performance gap may be partly attributable to differences in model architecture, training objective, and training data scale. 
Although Soft PER reduces absolute error rates for several languages, it does not remove cross-lingual performance disparities or change the overall model ranking. This indicates that some standard PER errors reflect acceptable phonemic variation, but that meaningful language-level differences remain even under relaxed matching.

The demographic evaluation also shows that the performance variation is not uniform across speaker identity characteristics. Gender-based differences are small and largely inconsistent across datasets and models, with one exception: ZIPA shows a small but statistically significant gender effect in WAXAL. Accent and ethnicity show the most robust effects: in SVC, ``Latino'' and ``Asian'' speakers have significantly higher PER than nearly every USA regional group across both models, while in EdAcc, ``Black'' speakers have significantly higher PER than ``South Asian'' speakers, and, for ZIPA specifically, than White speakers as well; ``Asian'' speakers, despite an elevated disparity from the overall mean, are not significantly different from any individual ethnicity group under pairwise testing. Age effects are more dataset-dependent, with a robust elevation for the youngest speakers in CORAAL, a modest but significant effect in SVC's 42--54 group, and no statistically significant effect in EdAcc despite a large descriptive gap for its smallest age group. Across demographic attributes, Soft PER lowers absolute error rates but generally preserves relative group-level patterns, suggesting that observed disparities are not fully explained by minor phonemic variation alone.


Overall, our findings show that IPA-based ASR systems exhibit demographic and language-level disparities in measured PER. However, these disparities are measured against G2P references, which may not capture legitimate pronunciation variation and can conflate differences in realized pronunciation with ASR error. Although phoneme-based representations provide a promising foundation for low-resource and multilingual speech technologies, their performance can still vary substantially between languages, accents, and demographic groups. These results highlight the need for more careful evaluation of phoneme-level ASR systems, richer demographic and sociophonetic benchmarks, and evaluation metrics that distinguish genuine recognition errors from acceptable pronunciation variation.

\newpage
\section*{Limitations}

The key limitation of phoneme-level evaluation is the reliance on auto-generated IPA ground truth, because of the high cost of specialized manual phonemic annotation. Unlike orthographic transcriptions, phonemic annotations depend on G2P conversion and forced alignment, which can introduce systematic noise and model bias in the ground truth. Because our pipeline converts written transcripts into IPA representations, the resulting “ground truth” often reflects standardized pronunciation, which can encode biases of what constitutes “correct” speech into our analysis \cite{vythelingum2017error, mackenzie2020assessing}. This issue will naturally increase the error rates for accented speech, dialectal variation, and non-canonical pronunciations, where no single canonical phonemic form may exist \cite{fatema2024ipatranscriptionbengalitexts}. Even with a softened PER, it may reflect annotation artifacts rather than true model error. Addressing this limitation may require future evaluation frameworks that utilize expert-validated phonemic annotations.

Our Soft PER hopes to partially mitigate sensitivity to annotation noise. While this acts as a necessary smoothing, it introduces a tradeoff between linguistic fidelity and metric standardization. Our definition of Soft PER may not be directly comparable to standard PER or WER, limiting cross-study comparability. Additionally, the equivalence mapping relies on English-prioritized relations when a surface phone appears across multiple languages. Although this provides a consistent normalization strategy, it may bias the metric toward English phonological interpretations and over-normalize distinctions that remain contrastive in other languages. Future work should explore language-adaptive mappings that better preserve cross-lingual phonological variation.

The analysis is further constrained by the demographic metadata available in the evaluated datasets. These categories do not capture finer sociophonetic variation such as code-switching, intra-speaker variability, or regional micro-dialects. In addition, certain demographic groups are underrepresented, which may limit the statistical robustness and generalizability of the results. Future work could address this by incorporating or creating a dataset with richer sociolinguistic annotations optimized for multilingual analysis to better understand the relationship among different dimensions.

Finally, while IPA provides a more fine-grained representation of speech than grapheme-based transcription, it is an abstract acoustic representation. There is ongoing debate in linguistics regarding the extent to which phonemic representations can fully capture gradient phonetic variation \cite{mallaband2024agreement, mitterer2018allophones}. Dialectal variation and allophonic realizations may not be fully represented within standardized IPA inventories, which may further influence both the accuracy of the ground truth and evaluation results \cite{fatema2024ipatranscriptionbengalitexts}. Hybrid evaluation frameworks in the future that jointly leverage phonemic representations and acoustic similarity measures could potentially provide a more accurate representation of ASR performance across diverse demographics.

\bibliography{reference}

\clearpage

\appendix

\section{Appendix}

\subsection{Dataset Details}\label{appendix:dataset}

Table~\ref{appendix:dataset_overview} summarizes the multilingual speech datasets used for cross-lingual evaluation, including the languages represented in each dataset and the dataset split used for analysis. Together, these datasets support evaluation across both high-resource and lower-resource languages.

Table~\ref{appendix:dataset_demographics} provides a detailed summary of the English-language datasets and the speaker-level demographic metadata available in each dataset. These metadata fields, including age, gender, ethnicity, accent, and dialectal region, are used to evaluate whether IPA-based ASR models exhibit systematic performance variation across demographic groups.

\begin{table*}[ht]
\centering
\caption{Dataset Metadata and Speech Characteristics}
\label{appendix:dataset_overview}
\renewcommand{\arraystretch}{1.25}

\resizebox{\textwidth}{!}{
\begin{tabular}{p{2.3cm} p{3.6cm} p{1.1cm} p{2.1cm} p{3.8cm} p{3.8cm}}
\hline
\textbf{Dataset} & \textbf{Languages / Varieties} & \textbf{Subset}$^\dagger$ & \textbf{Collection Period} & \textbf{Speech Style} & \textbf{Recording Environment} \\
\hline

Fleurs\_IPA & English, Hindi, Bangla, Panjabi, Tamil, Telugu & Test & 2020--2022 & Read speech (utterance-level) & Unedited original recordings \\
\hline
MediaSpeech & Arabic, French, Spanish, Turkish & N/A & 2020 & Read and spontaneous speech & YouTube broadcasts/channels \\
\hline
Waxal & Shona & Train & Jan.\ 2021--Mar.\ 2024 & Natural, spontaneous speech & Natural environments \\
\hline
EdAcc & English & Test & 2023 & Naturalistic, spontaneous interactions & Zoom recordings \\
\hline
SVC & English & Train & 2024 & Read speech & Clean, close-field environments \\
\hline
CORAAL & African American English & Train & 2023 & Sociolinguistic interviews & Unedited original recordings \\
\hline

\end{tabular}
} 

\vspace{1mm}
{\footnotesize $^\dagger$ Dataset partition used for bias evaluation.}

\end{table*}
\vspace{5mm}










\begin{table*}[ht]
\centering
\small
\caption{Dataset Languages and Demographic Attributes}
\label{appendix:dataset_demographics}
\renewcommand{\arraystretch}{1.25}
\begin{tabular}{p{3cm} p{4cm} p{7cm}}
\hline
\textbf{Dataset} & \textbf{Language(s)} & \textbf{Demographic Attributes} \\
\hline

WAXAL &
Shona &
gender\\
\hline
EdAcc &
English &
gender, year\_of\_birth, ethnicity, education, l1, l2, accent, english\_since, countries\_lived, lang\_home, lang\_friends \\
\hline
SVC Bias Assessment &
English &
age\_group, gender, dialectal\_region \\
\hline
CORAAL &
African American English &
city, se\_group, age\_group, gender \\

\hline
\end{tabular}
\end{table*}

\subsection{IPA Transcription Post-Processing}\label{appendix:ipa-normalizer}

To ensure that the evaluation focused on phonemic correspondence rather than superficial transcription differences, we apply the same normalization procedure to both the G2P-generated ground-truth IPA transcriptions and the model-generated outputs before computing PER and Soft PER as described in Table~\ref{tab:ipa-normalization}. The normalization rules are designed to reduce variation caused by stress notation, IPA symbol variants, tokenization conventions, and formatting artifacts, while preserving the underlying phone sequence as much as possible.

After applying these transformations, the normalized phone strings were re-tokenized into whitespace-separated phone sequences. This produced a consistent representation across reference and generated outputs, reducing the impact of orthographic IPA variants and tokenization artifacts on downstream evaluation.

\begin{table*}[h]
\centering
\small
\caption{Normalization rules applied to both ground-truth IPA transcriptions and model-generated outputs prior to evaluation.}
\renewcommand{\arraystretch}{1.25}
\begin{tabular}{p{0.25\linewidth} p{0.68\linewidth}}
\hline
\textbf{Category} & \textbf{Normalization behavior} \\
\hline
Stress marks &
Primary and secondary stress marks, such as \textipa{"} and \textipa{,}, were removed. \\
Length marks &
Length markers, including \textipa{:}, \textipa{;}, and the ASCII colon \texttt{:}, were removed. \\
\hline
Rhotics &
Rhotic consonants such as \textipa{\*r}, \textipa{r}, \textipa{K}, and \textipa{K} were mapped to \textipa{r}. Rhotic vowels such as \textipa{\textrhookschwa} and \textipa{\textrhookrevepsilon} were expanded to \textipa{@ r} and \textipa{\textepsilon r}, respectively. Rhoticity marks, such as \textipa{@\super{r}}, were also normalized to a vowel followed by \textipa{r}. \\
\hline
Diphthongs &
Joined diphthong symbols were tokenized as separate phone sequences. For example, \textipa{eI} was normalized to \textipa{e I}. \\
\hline
Affricates &
Joined and tie-barred affricates were tokenized as separate phones. For example, \textipa{tS}, \textipa{t\texttoptiebar S}, and \textipa{\textteshlig} were normalized to \textipa{t S}. \\
\hline
Token separators &
Whitespace, SentencePiece markers such as \texttt{\textunderscore}, hyphens, slashes, and similar separators were collapsed into phone-token boundaries. \\
\hline
Combining marks &
Separated combining marks were attached to the preceding phone. For example, \textipa{A \~{}} was normalized to \textipa{\~{A}}. \\
\hline
Symbol choices &
Common spelling variants were standardized. For example, ASCII \texttt{g} was normalized to IPA \textipa{g}, and dark-\textit{l} variants such as \textipa{l\super{\textgamma}}, \textipa{l\super{G}}, and separated \textipa{l \super{G}} were normalized to \textipa{\l}. \\
\hline
\end{tabular}
\label{tab:ipa-normalization}
\end{table*}

\subsection{Soft PER Details}\label{appendix:soft-per}

Table~\ref{appendix:tier_one} lists the transitive phonetic equivalence classes used in the first tier of Soft PER. These classes group phones that are treated as interchangeable during evaluation, allowing the metric to avoid penalizing substitutions that correspond to linguistically similar or allophonic variants. The Tier~1 mapping is applied universally across languages and is intended to capture broad phonemic variation that should not necessarily be counted as a recognition error.

The Tier~1 equivalence classes are constructed from two sources of phonetic similarity. First, language-specific allophonic mappings are derived from AlloVera, which links surface phones to canonical phonemes across multiple languages. Second, articulatory feature representations from PHOIBLE are used to identify canonical phonemes that differ by only one articulatory feature. These phones are merged into shared transitive classes, producing a mapping that spans 62 equivalence classes and 254 phones. During Soft PER computation, substitutions within the same Tier~1 class receive zero penalty, allowing the metric to better distinguish substantial recognition errors from acceptable phonetic variation.

Table~\ref{appendix:tier_two} lists the language-specific non-transitive similarity pairs used in the second tier of Soft PER. Unlike the Tier~1 classes, these pairs are not merged into broader equivalence classes. Instead, they capture cases where the same surface phone corresponds to different canonical phonemes across languages. This allows Soft PER to recognize language-specific pronunciation overlap without incorrectly making all connected phonemes globally interchangeable.

The Tier~2 mapping is also derived from AlloVera. When a surface phone maps to different canonical phonemes in different languages, the corresponding canonical phonemes are recorded as a direct similarity pair. During evaluation, these pairs are applied only when relevant to the target language. Substitutions covered by Tier~2 receive zero penalty, but the relation remains non-transitive; for example, if two separate language-specific mappings relate one phone to two different phones, Soft PER does not automatically treat those two phones as equivalent to each other. This design preserves language-specific phonological distinctions while still reducing penalties for acceptable cross-lingual or allophonic variation.

\begin{table*}[ht]
\centering
\small
\caption{Phonetic Transitive Equivalence Classes (Tier 1).}
\label{appendix:tier_one}
\begin{tabular}{c p{0.78\textwidth}}
\toprule
\textbf{Index} & \textbf{Representations} \\
\midrule
0 & \textipa{b', p'} \\
1 & \textipa{b, \textlowering{b}, \textsubring{b}} \\
2 & \textipa{dZ, \t{dZ}} \\
3 & \textipa{d, \textsubring{d}} \\
4 & \textipa{e, \textlowering{e}} \\
5 & \textipa{k\super{w}', k'\super{w}} \\
6 & \textipa{k, \textcorner{\textsubplus{k}}} \\
7 & \textipa{o, \textlowering{o}, \textsubring{o}} \\
8 & \textipa{pf, \t{pf}} \\
9 & \textipa{ts, \t{ts}} \\
10 & \textipa{t\textctc, \t{t\textctc}} \\
11 & \textipa{tS, \t{tS}} \\
12 & \textipa{t\super{w}', t'\super{w}} \\
13 & \textipa{t, \textcorner{\textsubbridge{t}}} \\
14 & \textipa{v, \textlowering{v}} \\
15 & \textipa{x\super{j}, \textsubplus{x}\super{j}} \\
16 & \textipa{N, \textsubplus{N}} \\
17 & \textipa{A, \textsubbar{A}} \\
18 & \textipa{\textctc, \textctc:} \\
19 & \textipa{g, \textsubring{g}} \\
20 & \textipa{r\super{j}, \textfishhookr\super{j}} \\
21 & \textipa{\textlowering{\textbaru}, Y} \\
\bottomrule
\end{tabular}
\end{table*}

\begin{table*}[t]
\centering
\scriptsize
\caption{Language Specific Non-Transitive Pairs (Tier 2).}
\label{appendix:tier_two}
\begin{tabular}{llllll}
\toprule
\textbf{Phoneme 1} & \textbf{Phoneme 2} & \textbf{Language} &
\textbf{Phoneme 1} & \textbf{Phoneme 2} & \textbf{Language} \\
\midrule

\ipa{a} & \ipa{a:} & \lang{jpn, tur} &
\ipa{a} & \ipa{\ae} & \lang{deu, eng} \\

\ipa{a} & \ipa{A} & \lang{cmn, eng} &
\ipa{a} & \ipa{O} & \lang{eng, jav} \\

\ipa{a} & \ipa{@} & \lang{eng, rus} &
\ipa{a} & \ipa{E} & \lang{cmn, eng} \\

\ipa{a} & \ipa{V} & \lang{eng, rus} &
\ipa{a:} & \ipa{A} & \lang{deu, eng} \\

\ipa{b} & \ipa{b\super j} & \lang{jpn, rus} &
\ipa{b} & \ipa{p} & \lang{amh, eng, kaz, rus, spa} \\

\ipa{b} & \ipa{v} & \lang{spa, tur} &
\ipa{b\super j} & \ipa{p\super j} & \lang{rus} \\

\ipa{d} & \ipa{dZ} & \lang{ita, tgl} &
\ipa{d} & \ipa{t} & \lang{eng, kaz, rus, vie} \\

\ipa{d} & \ipa{D} & \lang{eng, spa} &
\ipa{d} & \ipa{4} & \lang{spa, tgl} \\

\ipa{d\super j} & \ipa{t\super j} & \lang{rus} &
\ipa{dZ} & \ipa{ts} & \lang{ita, tgl} \\

\ipa{dZ} & \ipa{Z} & \lang{amh, eng, ita, tgl} &
\ipa{e} & \ipa{e:} & \lang{deu, eng, jpn, tur} \\

\ipa{e} & \ipa{i} & \lang{eng, tgl} &
\ipa{e} & \ipa{\ae} & \lang{eng, tur} \\

\ipa{e} & \ipa{E} & \lang{deu, eng} &
\ipa{e} & \ipa{I} & \lang{eng, tgl} \\

\ipa{e} & \ipa{V} & \lang{cmn, eng} &
\ipa{h} & \ipa{r} & \lang{fra, kaz} \\

\ipa{h} & \ipa{x} & \lang{deu, tur} &
\ipa{h} & \ipa{C} & \lang{deu, jpn} \\

\ipa{i} & \ipa{i:} & \lang{deu, eng, jpn, tur} &
\ipa{i} & \ipa{E} & \lang{eng, tgl} \\

\ipa{i} & \ipa{I} & \lang{deu, eng} &
\ipa{j} & \ipa{y} & \lang{eng, jpn} \\

\ipa{j} & \ipa{C} & \lang{deu, rus, spa} &
\ipa{j} & \ipa{H} & \lang{cmn, fra} \\

\ipa{j} & \ipa{I} & \lang{eng} &
\ipa{j} & \ipa{L} & \lang{deu, spa} \\

\ipa{k} & \ipa{q} & \lang{eng, jav} &
\ipa{k} & \ipa{x} & \lang{deu, tgl} \\

\ipa{k} & \ipa{g} & \lang{amh, deu, eng, kaz, rus} &
\ipa{k} & \ipa{g\super j} & \lang{rus} \\

\ipa{m} & \ipa{m\super j} & \lang{jpn, rus} &
\ipa{m} & \ipa{n} & \lang{eng, ita} \\

\ipa{m} & \ipa{N} & \lang{eng, ita, tgl} &
\ipa{m} & \ipa{\textltailn} & \lang{eng, fra, ita} \\

\ipa{n} & \ipa{N} & \lang{eng, spa, tgl} &
\ipa{n} & \ipa{\textltailn} & \lang{eng, fra, ita} \\

\ipa{o} & \ipa{o:} & \lang{deu, eng} &
\ipa{o} & \ipa{u} & \lang{eng, tgl} \\

\ipa{o} & \ipa{O} & \lang{eng} &
\ipa{o} & \ipa{U} & \lang{eng, tgl} \\

\ipa{p} & \ipa{p\super j} & \lang{jpn, rus} &
\ipa{p} & \ipa{q} & \lang{jpn, vie} \\

\ipa{q} & \ipa{r} & \lang{fra, kaz} &
\ipa{q} & \ipa{s} & \lang{eng, jpn} \\

\ipa{r} & \ipa{x} & \lang{deu, fra, kaz, spa} &
\ipa{r} & \ipa{\textturnr} & \lang{eng, spa} \\

\ipa{r} & \ipa{4} & \lang{jpn, spa, tgl} &
\ipa{r} & \ipa{S} & \lang{eng, kaz} \\

\ipa{r} & \ipa{Z} & \lang{eng, kaz} &
\ipa{s} & \ipa{z} & \lang{amh, deu, eng, kaz} \\

\ipa{s} & \ipa{S} & \lang{eng, tgl} &
\ipa{s} & \ipa{T} & \lang{eng, spa} \\

\ipa{t} & \ipa{tS} & \lang{spa, tgl} &
\ipa{ts} & \ipa{tS} & \lang{spa, tgl} \\

\ipa{ts} & \ipa{z} & \lang{ita, jpn} &
\ipa{u} & \ipa{u:} & \lang{deu, eng, jpn, tur} \\

\ipa{u} & \ipa{w} & \lang{eng, ita} &
\ipa{u} & \ipa{1} & \lang{jpn, rus} \\

\ipa{u} & \ipa{U} & \lang{deu, eng} &
\ipa{w} & \ipa{U} & \lang{eng} \\

\ipa{x} & \ipa{g} & \lang{rus, spa} &
\ipa{y} & \ipa{y:} & \lang{deu, fra} \\

\ipa{\o} & \ipa{\o:} & \lang{deu, fra} &
\ipa{\o} & \ipa{\oe} & \lang{deu, fra} \\

\ipa{N} & \ipa{\textltailn} & \lang{eng, ita} &
\ipa{O} & \ipa{@} & \lang{amh, eng} \\

\ipa{@} & \ipa{E} & \lang{amh, eng} &
\ipa{g} & \ipa{\textturnmrleg} & \lang{tgl, tur} \\  

\ipa{1} & \ipa{I} & \lang{amh, eng} &
\ipa{1} & \ipa{M} & \lang{rus, vie} \\  

\ipa{1} & \ipa{U} & \lang{amh, eng} &
\ipa{\textturnr} & \ipa{4} & \lang{eng, tgl} \\  

\ipa{S} & \ipa{Z} & \lang{eng, kaz} &
\ipa{L} & \ipa{Z} & \lang{eng, spa} \\

\bottomrule
\end{tabular}
\end{table*}

\subsection{Pairwise Welch's $t$-test Details}
\label{appendix:welch-ttest}

Tables~\ref{tab:sig-gender-all}--\ref{tab:sig-accent-svc} report pairwise comparisons of speaker-level PER between demographic groups within each dataset, for the demographic attributes examined in Section~\ref{sec:demographic-performance}: gender, age, ethnicity, and accent/dialect. For every pair of groups within a demographic attribute, we compute the difference in group-mean PER (row minus column) and test the difference using a two-sided Welch's $t$-test. This is reported separately for ZIPA and WhisperIPA, under both standard PER and Soft PER.

In each table, the reported value is the point-estimate difference in speaker-level mean PER between the row groups and column groups. A positive value indicates that the row group has a higher (worse) mean PER than the column group.

Group-level means and their 95\% confidence intervals, computed using bias-corrected and accelerated (BCa) bootstrap with 10{,}000 resamples (percentile intervals when a group contains fewer than 10 speakers or when the BCa procedure does not converge), are the same estimates used for the error bars in the demographic performance figures in Section~\ref{sec:demographic-performance}.

\begin{table*}[]
    \centering
\renewcommand{\arraystretch}{1.25}
\small
\caption{Gender bias analysis across all four demographic datasets. Values represent the difference in speaker-level mean PER between the row and column groups (row $-$ column). Statistically significant difference in
mean between two groups via Welch’s $t$-test is shown as: $*$: $p<0.05$; $**$: $p<0.01$; $***$: $p<0.001$. Positive values indicate higher PER for the first-listed group.}
    \begin{tabular}{l l c c c c}
        \textbf{Dataset} & \textbf{Groups} & \textbf{ZIPA PER} & \textbf{ZIPA Soft PER} & \textbf{Whisper PER} & \textbf{Whisper Soft PER} \\ \hline
        SVC & Female vs Male & +0.001 & +0.001 & -0.001 & +0.000 \\
        EdAcc & Female vs Male & -0.017 & -0.012 & +0.028 & +0.028 \\
        WAXAL & Female vs Male & \textbf{+0.067**} & \textbf{+0.067**} & -0.149 & -0.149 \\
        CORAAL & Female vs Male & -0.025 & -0.019 & +0.011 & +0.015 \\
        \hline
    \end{tabular}
    \label{tab:sig-gender-all}
\end{table*}
\vspace{5mm}

\begin{table*}[]
    \centering
    \small
\renewcommand{\arraystretch}{1.25}
\caption{Ethnicity bias analysis on the EdAcc dataset. Values are the difference in speaker-level mean PER between the row and column groups (row $-$ column). Statistically significant difference in
mean between two groups via Welch’s $t$-test is shown as: $*$: $p<0.05$; $**$: $p<0.01$; $***$: $p<0.001$. Positive values indicate higher PER for the first-listed group.}
    \begin{tabular}{llccc}
         & & Black & South Asian & White \\ \hline
        ZIPA PER          & Asian              & -0.038 & +0.044 & +0.036 \\
        ZIPA Soft PER     &                    & -0.023 & +0.039 & +0.031 \\
        Whisper PER       &                    & -0.010 & +0.147 & +0.075 \\
        Whisper Soft PER  &                    & +0.004 & +0.148 & +0.071 \\
        \hline
        ZIPA PER          & Black              &  & \textbf{+0.082*} & \textbf{+0.074*} \\
        ZIPA Soft PER     &                    &  & +0.062 & \textbf{+0.054*} \\
        Whisper PER       &                    &  & \textbf{+0.157**} & +0.085 \\
        Whisper Soft PER  &                    &  & \textbf{+0.144*} & +0.067 \\
        \hline
        ZIPA PER          & South Asian        &  &  & -0.008 \\
        ZIPA Soft PER     &                    &  &  & -0.008 \\
        Whisper PER       &                    &  &  & -0.072 \\
        Whisper Soft PER  &                    &  &  & -0.077 \\
        \hline
    \end{tabular}
    \label{tab:sig-ethnicity-edacc}
\end{table*}
\vspace{5mm}

\begin{table*}[]
    \centering
    \small
    \renewcommand{\arraystretch}{1.25}
    \caption{Age-group bias analysis on the SVC dataset. Values are the difference in speaker-level mean PER between the row and column groups (row $-$ column). Statistically significant difference in mean between two groups via Welch’s $t$-test is shown as: $*$: $p<0.05$; $**$: $p<0.01$; $***$: $p<0.001$. Positive values indicate higher PER for the first-listed group.}
    \begin{tabular}{llcccc}
         & & 17-28 & 29-41 & 42-54 & 55-100 \\ \hline
        ZIPA PER          & 9-16               & -0.005 & +0.001 & \textbf{+0.014*} & +0.007 \\
        ZIPA Soft PER     &                    & -0.004 & +0.002 & \textbf{+0.011**} & +0.005 \\
        Whisper PER       &                    & -0.004 & +0.005 & +0.019 & -0.006 \\
        Whisper Soft PER  &                    & +0.001 & +0.008 & \textbf{+0.016*} & -0.003 \\
        \hline
        ZIPA PER          & 17-28              &  & +0.006 & \textbf{+0.019**} & +0.012 \\
        ZIPA Soft PER     &                    &  & +0.006 & \textbf{+0.015**} & +0.009 \\
        Whisper PER       &                    &  & +0.009 & \textbf{+0.023*} & -0.002 \\
        Whisper Soft PER  &                    &  & +0.007 & \textbf{+0.015*} & -0.004 \\
        \hline
        ZIPA PER          & 29-41              &  &  & +0.013 & +0.006 \\
        ZIPA Soft PER     &                    &  &  & +0.009 & +0.003 \\
        Whisper PER       &                    &  &  & +0.014 & -0.011 \\
        Whisper Soft PER  &                    &  &  & +0.008 & -0.011 \\
        \hline
        ZIPA PER          & 42-54              &  &  &  & -0.007 \\
        ZIPA Soft PER     &                    &  &  &  & -0.006 \\
        Whisper PER       &                    &  &  &  & -0.025 \\
        Whisper Soft PER  &                    &  &  &  & -0.019 \\
        \hline
    \end{tabular}
    \label{tab:sig-age-svc}
\end{table*}
\vspace{5mm}

\begin{table*}[]
    \centering
    \small
    \renewcommand{\arraystretch}{1.25}
    \caption{Age-group bias analysis on the EdAcc dataset. Values are the difference in speaker-level mean PER between the row and column groups (row $-$ column). Statistically significant difference in mean between two groups via Welch’s $t$-test is shown as: $*$: $p<0.05$; $**$: $p<0.01$; $***$: $p<0.001$. Positive values indicate higher PER for the first-listed group.}
    \begin{tabular}{llcc}
         & & 30--44 & 45--59 \\ \hline
        ZIPA PER          & 18--29             & +0.006 & +0.024 \\
        ZIPA Soft PER     &                    & +0.008 & +0.021 \\
        Whisper PER       &                    & +0.079 & -0.063 \\
        Whisper Soft PER  &                    & +0.087 & -0.057 \\
        \hline
        ZIPA PER          & 30--44             &  & +0.018 \\
        ZIPA Soft PER     &                    &  & +0.013 \\
        Whisper PER       &                    &  & -0.142 \\
        Whisper Soft PER  &                    &  & -0.144 \\
        \hline
    \end{tabular}
    \label{tab:sig-age-edacc}
\end{table*}
\vspace{5mm}

\begin{table*}[]
    \centering
    \small
 
    \renewcommand{\arraystretch}{1.25}
    \caption{Age-group bias analysis on the CORAAL dataset. Values are the difference in speaker-level mean PER between the row and column groups (row $-$ column). Statistically significant difference in mean between two groups via Welch’s $t$-test is shown as: $*$: $p<0.05$; $**$: $p<0.01$; $***$: $p<0.001$. Positive values indicate higher PER for the first-listed group.}
    \begin{tabular}{llcccc}
         & & 0-29 & 20-29 & 30-50 & 51+ \\ \hline
        ZIPA PER          & 0-19               & \textbf{+0.094**} & \textbf{+0.092***} & \textbf{+0.099***} & \textbf{+0.085***} \\
        ZIPA Soft PER     &                    & \textbf{+0.073**} & \textbf{+0.070***} & \textbf{+0.077***} & \textbf{+0.067***} \\
        Whisper PER       &                    & +0.104 & \textbf{+0.159***} & \textbf{+0.131***} & \textbf{+0.122***} \\
        Whisper Soft PER  &                    & +0.096 & \textbf{+0.152***} & \textbf{+0.120**} & \textbf{+0.112**} \\
        \hline
        ZIPA PER          & 0-29               &  & -0.002 & +0.005 & -0.009 \\
        ZIPA Soft PER     &                    &  & -0.003 & +0.004 & -0.006 \\
        Whisper PER       &                    &  & +0.055 & +0.027 & +0.018 \\
        Whisper Soft PER  &                    &  & +0.056 & +0.024 & +0.016 \\
        \hline
        ZIPA PER          & 20-29              &  &  & +0.007 & -0.007 \\
        ZIPA Soft PER     &                    &  &  & +0.007 & -0.003 \\
        Whisper PER       &                    &  &  & -0.028 & -0.037 \\
        Whisper Soft PER  &                    &  &  & -0.032 & -0.040 \\
        \hline
        ZIPA PER          & 30-50              &  &  &  & -0.014 \\
        ZIPA Soft PER     &                    &  &  &  & -0.010 \\
        Whisper PER       &                    &  &  &  & -0.009 \\
        Whisper Soft PER  &                    &  &  &  & -0.008 \\
        \hline
    \end{tabular}
    \label{tab:sig-age-coraal}
\end{table*}
\vspace{5mm}

\begin{table*}[ht]
    \centering
    \renewcommand{\arraystretch}{1.25}
    \caption{Accent/dialect bias analysis on the SVC dataset. Region names abbreviated (Inland-N.=USA Inland-North, Mid-Atl.=USA Mid-Atlantic, New Eng.=USA New England). Values are the difference in speaker-level mean PER between the row and column groups (row $-$ column). Statistically significant difference in mean between two groups via Welch’s $t$-test is shown as: $*$: $p<0.05$; $**$: $p<0.01$; $***$: $p<0.001$. Positive values indicate higher PER for the first-listed group.}
    \small
    \resizebox{\textwidth}{!}{
    \begin{tabular}{llccccccc}
         & & Latino & Inland-N. & Mid-Atl. & Midland & New Eng. & Southern & Western \\ \hline
        ZIPA PER          & Asian              & \textbf{-0.033**} & \textbf{+0.038***} & \textbf{+0.039***} & \textbf{+0.035***} & \textbf{+0.040***} & \textbf{+0.029**} & \textbf{+0.036***} \\
        ZIPA Soft PER     &                    & \textbf{-0.021*} & \textbf{+0.023***} & \textbf{+0.025***} & \textbf{+0.022***} & \textbf{+0.026***} & \textbf{+0.018**} & \textbf{+0.024***} \\
        Whisper PER       &                    & -0.020 & \textbf{+0.058***} & \textbf{+0.066***} & \textbf{+0.058***} & \textbf{+0.064***} & \textbf{+0.049***} & \textbf{+0.058***} \\
        Whisper Soft PER  &                    & -0.007 & \textbf{+0.035**} & \textbf{+0.041***} & \textbf{+0.035**} & \textbf{+0.039**} & \textbf{+0.029**} & \textbf{+0.035**} \\
        \hline
        ZIPA PER          & Latino             &  & \textbf{+0.071***} & \textbf{+0.072***} & \textbf{+0.068***} & \textbf{+0.073***} & \textbf{+0.062***} & \textbf{+0.069***} \\
        ZIPA Soft PER     &                    &  & \textbf{+0.044***} & \textbf{+0.046***} & \textbf{+0.043***} & \textbf{+0.047***} & \textbf{+0.039***} & \textbf{+0.045***} \\
        Whisper PER       &                    &  & \textbf{+0.078***} & \textbf{+0.086***} & \textbf{+0.078***} & \textbf{+0.084***} & \textbf{+0.069***} & \textbf{+0.078***} \\
        Whisper Soft PER  &                    &  & \textbf{+0.042***} & \textbf{+0.048***} & \textbf{+0.042***} & \textbf{+0.046***} & \textbf{+0.036***} & \textbf{+0.042***} \\
        \hline
        ZIPA PER          & Inland-N.          &  &  & +0.001 & -0.003 & +0.002 & -0.009 & -0.002 \\
        ZIPA Soft PER     &                    &  &  & +0.002 & -0.001 & +0.003 & \textbf{-0.005*} & +0.001 \\
        Whisper PER       &                    &  &  & +0.008 & +0.000 & +0.006 & -0.009 & +0.000 \\
        Whisper Soft PER  &                    &  &  & +0.006 & +0.000 & +0.004 & -0.006 & +0.000 \\
        \hline
        ZIPA PER          & Mid-Atl.           &  &  &  & -0.004 & +0.001 & -0.010 & -0.003 \\
        ZIPA Soft PER     &                    &  &  &  & -0.003 & +0.001 & \textbf{-0.007*} & -0.001 \\
        Whisper PER       &                    &  &  &  & -0.008 & -0.002 & \textbf{-0.017*} & -0.008 \\
        Whisper Soft PER  &                    &  &  &  & -0.006 & -0.002 & -0.012 & -0.006 \\
        \hline
        ZIPA PER          & Midland            &  &  &  &  & +0.005 & -0.006 & +0.001 \\
        ZIPA Soft PER     &                    &  &  &  &  & +0.004 & -0.004 & +0.002 \\
        Whisper PER       &                    &  &  &  &  & +0.006 & -0.009 & +0.000 \\
        Whisper Soft PER  &                    &  &  &  &  & +0.004 & -0.006 & +0.000 \\
        \hline
        ZIPA PER          & New Eng.           &  &  &  &  &  & -0.011 & -0.004 \\
        ZIPA Soft PER     &                    &  &  &  &  &  & \textbf{-0.008*} & -0.002 \\
        Whisper PER       &                    &  &  &  &  &  & -0.015 & -0.006 \\
        Whisper Soft PER  &                    &  &  &  &  &  & -0.010 & -0.004 \\
        \hline
        ZIPA PER          & Southern           &  &  &  &  &  &  & +0.007 \\
        ZIPA Soft PER     &                    &  &  &  &  &  &  & +0.006 \\
        Whisper PER       &                    &  &  &  &  &  &  & +0.009 \\
        Whisper Soft PER  &                    &  &  &  &  &  &  & +0.006 \\
        \hline
    \end{tabular}
    } 
    \label{tab:sig-accent-svc}
\end{table*}

\end{document}